\pdfoutput=1

\documentclass[11pt]{article}

\usepackage[preprint]{acl}

\usepackage{times}
\usepackage{latexsym}
\usepackage{hyperref}

\usepackage[T1]{fontenc}

\usepackage[utf8]{inputenc}

\usepackage{microtype}

\usepackage{inconsolata}

\usepackage{graphicx}
\usepackage{multirow}
\usepackage{subfigure}
\usepackage{subcaption}
\usepackage{tabularx}
\usepackage{booktabs} 
\usepackage[table]{xcolor}
\usepackage{ragged2e}
\usepackage[vlined,ruled]{algorithm2e}
\usepackage{amsmath}
\usepackage{amssymb}
\usepackage{mathtools}
\usepackage{amsthm}
\usepackage{parskip} 
\usepackage{tcolorbox}
\theoremstyle{plain}
\newtheorem{example}{Example}

\theoremstyle{remark}

\newcommand{\shrink}[1]{}

\usepackage[textsize=tiny]{todonotes}

\title{\textsc{FactCorrector}: A Graph-Inspired Approach \\to Long-Form Factuality Correction of Large Language Models}

\author{
  \textbf{Javier Carnerero-Cano\textsuperscript{1}},
  \textbf{Massimiliano Pronesti\textsuperscript{1}},
    \textbf{Radu Marinescu\textsuperscript{1}},
  \textbf{Tigran Tchrakian\textsuperscript{1}}, \\
  \textbf{James Barry\textsuperscript{1}},
  \textbf{Jasmina Gajcin\textsuperscript{1}},
  \textbf{Yufang Hou\textsuperscript{1,2}},
  \textbf{Alessandra Pascale\textsuperscript{1}},
  \textbf{Elizabeth Daly\textsuperscript{1}}\\
  \\
  \textsuperscript{1}IBM Research Europe - Ireland\\
  \textsuperscript{2}IT:U - Interdisciplinary Transformation University Austria\\
  {\small\url{{javier.cano,massimiliano.pronesti,james.barry,jasmina.gajcin2,yufang.hou1}@ibm.com}}\\
{\small\url{{radu.marinescu,tigran,apascale,elizabeth.daly}@ie.ibm.com}}
}

\begin{document}
\newcommand{\bA}{\mathbf{A}}
\newcommand{\bB}{\mathbf{B}}
\newcommand{\bC}{\mathbf{C}}
\newcommand{\bD}{\mathbf{D}}
\newcommand{\bE}{\mathbf{E}}
\newcommand{\bF}{\mathbf{F}}
\newcommand{\bG}{\mathbf{G}}
\newcommand{\bH}{\mathbf{H}}
\newcommand{\bI}{\mathbf{I}}
\newcommand{\bJ}{\mathbf{J}}
\newcommand{\bK}{\mathbf{K}}
\newcommand{\bL}{\mathbf{L}}
\newcommand{\bM}{\mathbf{M}}
\newcommand{\bN}{\mathbf{N}}
\newcommand{\bO}{\mathbf{O}}
\newcommand{\bP}{\mathbf{P}}
\newcommand{\bQ}{\mathbf{Q}}
\newcommand{\bR}{\mathbf{R}}
\newcommand{\bS}{\mathbf{S}}
\newcommand{\bT}{\mathbf{T}}
\newcommand{\bU}{\mathbf{U}}
\newcommand{\bV}{\mathbf{V}}
\newcommand{\bW}{\mathbf{W}}
\newcommand{\bX}{\mathbf{X}}
\newcommand{\bY}{\mathbf{Y}}
\newcommand{\bZ}{\mathbf{Z}}

\newcommand{\ba}{\mathbf{a}}
\newcommand{\bb}{\mathbf{b}}
\newcommand{\bc}{\mathbf{c}}
\newcommand{\bd}{\mathbf{d}}
\newcommand{\be}{\mathbf{e}}
\newcommand{\bbf}{\mathbf{f}}
\newcommand{\bg}{\mathbf{g}}
\newcommand{\bh}{\mathbf{h}}
\newcommand{\bi}{\mathbf{i}}
\newcommand{\bj}{\mathbf{j}}
\newcommand{\bk}{\mathbf{k}}
\newcommand{\bl}{\mathbf{l}}
\newcommand{\bm}{\mathbf{m}}
\newcommand{\bn}{\mathbf{n}}
\newcommand{\bo}{\mathbf{o}}
\newcommand{\bp}{\mathbf{p}}
\newcommand{\bq}{\mathbf{q}}
\newcommand{\br}{\mathbf{r}}
\newcommand{\bs}{\mathbf{s}}
\newcommand{\by}{\mathbf{y}}
\newcommand{\bu}{\mathbf{u}}
\newcommand{\bv}{\mathbf{v}}
\newcommand{\bw}{\mathbf{w}}
\newcommand{\bx}{\mathbf{x}}
\newcommand{\bby}{\mathbf{y}}
\newcommand{\bz}{\mathbf{z}}

\newcommand{\cA}{\mathcal{A}}
\newcommand{\cB}{\mathcal{B}}
\newcommand{\cC}{\mathcal{C}}
\newcommand{\cD}{\mathcal{D}}
\newcommand{\cE}{\mathcal{E}}
\newcommand{\cF}{\mathcal{F}}
\newcommand{\cG}{\mathcal{G}}
\newcommand{\cH}{\mathcal{H}}
\newcommand{\cI}{\mathcal{I}}
\newcommand{\cJ}{\mathcal{J}}
\newcommand{\cK}{\mathcal{K}}
\newcommand{\cL}{\mathcal{L}}
\newcommand{\cM}{\mathcal{M}}
\newcommand{\cN}{\mathcal{N}}
\newcommand{\cO}{\mathcal{O}}
\newcommand{\cP}{\mathcal{P}}
\newcommand{\cQ}{\mathcal{Q}}
\newcommand{\cR}{\mathcal{R}}
\newcommand{\cS}{\mathcal{S}}
\newcommand{\cT}{\mathcal{T}}
\newcommand{\cU}{\mathcal{U}}
\newcommand{\cV}{\mathcal{V}}
\newcommand{\cW}{\mathcal{W}}
\newcommand{\cX}{\mathcal{X}}
\newcommand{\cY}{\mathcal{Y}}
\newcommand{\cZ}{\mathcal{Z}}

\def\inmath#1{\relax\ifmmode#1\else$#1$\fi}


\outer\def\operators#1{\begingroup
        \def\\##1##2{\gdef##1{{\mathop{\rm ##2}}}}\relax
        \\#1\endgroup
}

\outer\def\variables#1{\begingroup
        \def\\##1##2{\gdef##1{{\mathord{\it ##2}}}}\relax
        \\#1\endgroup
}

\maketitle
\begin{abstract}
    Large language models (LLMs) are widely used in knowledge-intensive applications but often generate factually incorrect responses. A promising approach to rectify these flaws is correcting LLMs using feedback. Therefore, in this paper, we introduce \textsc{FactCorrector}, a new post-hoc correction method that adapts across domains without retraining and leverages structured feedback about the factuality of the original response to generate a correction. To support rigorous evaluations of factuality correction methods, we also develop the \textsc{Veli5} benchmark, a novel dataset containing systematically injected factual errors and ground-truth corrections. Experiments on \textsc{Veli5} and several popular long-form factuality datasets show that the \textsc{FactCorrector} approach significantly improves factual precision while preserving relevance, outperforming strong baselines. We release
our code at \href{https://ibm.biz/factcorrector}{https://ibm.biz/factcorrector}.
\end{abstract}

\section{Introduction}
\label{sec-intro}

Large Language Models (LLMs) have demonstrated remarkable capabilities in generating coherent and contextually relevant text across diverse domains \cite{LLMFewShortLearner,chowdhery2022palm}. However, their tendency to produce factually incorrect or hallucinated content remains a significant barrier to safe and reliable deployment in real-world applications \cite{zhang2023siren,sahoo2024hallu,huang2025hallu}, where knowledge conflicts \cite{xu2024knowledge, hou2024wiki} are also common. Factuality is critical in high-stakes domains such as healthcare or finance, where even minor inaccuracies can lead to harmful decisions or financial loss \cite{tonmoy2024hallu}.

Prior research has explored several strategies to mitigate factual errors in LLM outputs. Specifically, training-time correction methods optimize the model behavior to avoid factual errors during learning by incorporating human or automated feedback. Approaches include direct optimization with human feedback \cite{glaese2022}, reward modeling and RLHF \cite{ouyang2022rlhf,zhong2025rewards}, and self-training with automated signals \cite{dubois2024alpacafarm}. However, these strategies are often infeasible for closed-source or large scale models \cite{pan2024survey}. Generation-time correction methods refine LLM outputs during decoding by leveraging critic models for guidance. Approaches such as generate-then-rank sample multiple candidate responses and select the best using a ranking model \cite{weng2023self}, while feedback-guided decoding integrates step-level feedback to steer generation in real time \cite{yao2023tot}. The effectiveness of these methods hinges on the critic’s ability to provide high-quality intermediate feedback. In contrast, post-hoc correction methods evaluate the final LLM response to construct the feedback that is subsequently used to refine and correct the response. The feedback can be provided through self-critique \cite{shinn2023reflexion}, external critics (e.g., RAC \cite{li2024rac}, CRITIC \cite{gou2024critic}) or agentic debate \cite{cohen2023lm}. However, current approaches are limited in that they accept feedback in a simple text format, which may miss the complex structure of relationships between statements to verify and their related evidence.

\paragraph{Contribution.} In this paper, we present \textsc{FactCorrector}, a novel post-hoc method for long-form factuality correction that leverages structured feedback to produce factually accurate responses. The approach begins by applying a critic model to the generated output to assess its factuality. This critic, built upon \textsc{FactReasoner}—a recent long-form factuality assessor \cite{marinescu2025fr}—decomposes the response into atomic units, each representing a single fact or claim. For each atom, it retrieves supporting or contradicting evidence from external sources and estimates its truthfulness using a graphical model that captures entailment and contradiction relationships between atoms and contexts. The resulting feedback identifies the false atoms along with the evidence that refutes them. Subsequently, a refinement model integrates this feedback with the original response into a structured prompt to generate a corrected version of the initial response, ensuring that the model focuses on the parts of the response that were flagged as incorrect by the critic. The approach can be applied in a single pass or iteratively until the desired factuality accuracy is achieved.

We also introduce the \textsc{Veli5} benchmark, a dataset specifically designed to enable rigorous, large-scale evaluation of long-form factuality correction methods and to support the training of specialized models for correcting LLM-generated responses. \textsc{Veli5} builds upon the widely used ELI5 dataset \cite{fan-etal-2019-eli5}, comprising curated pairs of human-authored questions and answers that have been explicitly verified and corrected for factual inaccuracies using \textsc{FactCorrector}.

Finally, we present an extensive empirical evaluation with our \textsc{Veli5} and several popular long-form factuality datasets using a wide range of open-source LLMs. The results show clearly that \textsc{FactCorrector} consistently delivers the most reliable improvements in factuality across all datasets thus outperforming strong baselines and demonstrating strong generalization across models.

\section{Related Work}
\label{sec-related}

Although LLMs have achieved remarkable performance across a wide range of tasks, they still struggle with hallucinations, unfaithful reasoning, or the propagation of bias and toxicity. In recent years, several techniques have been developed to address these issues by leveraging automated feedback -- either generated by the LLM itself or provided by external systems -- to refine and \emph{correct} model outputs \cite{pan2024survey}. These correction strategies can be broadly categorized into three types: training-time, generation-time, and post-hoc approaches.

\emph{Training-time correction} strategies aim to improve model behavior during the training phase by incorporating human feedback, reward models, or automated feedback. Common approaches include direct optimization using human feedback e.g., \cite{glaese2022,scheurer2024,chen2024,liu2023,gao2023}, reward modeling and reinforcement learning from human feedback (RLHF) e.g., \cite{ouyang2022rlhf,bai2022,ganguli2023}, and self-training with automated feedback e.g.,  \cite{zelikman2022star,bai2022,dubois2024alpacafarm}. However, these methods are often impractical for closed-source models or extremely large-scale models with billions of parameters.

\emph{Generation-time correction} methods, such as generate-then-rank and feedback-guided decoding, aim to refine the output of large language models (LLMs) during the generation process itself. In generate-then-rank approaches, a large pool of candidate outputs is first sampled, and then ranked using a secondary critic model to select the most appropriate response \cite{weng2023self, he2022rr, ni2023lever, chen2022codet}. In contrast, feedback-guided decoding incorporates a step-level critic model that provides real-time feedback during generation, enabling more fine-grained control. This strategy underpins methods like Tree-of-Thought \cite{yao2023tot}, GRACE \cite{khalifa2023grace}, and RAP \cite{hao2023rap}. These approaches primarily differ in the type of critic model employed—ranging from reward models trained with human feedback, to verifiers, external metrics or external knowledge sources.

The effectiveness of generation-time correction depends on the critic model’s ability to deliver high-quality feedback on intermediate outputs. In contrast, \emph{post-hoc correction} methods apply both the critic and refinement models after the full output is generated, enabling richer and more varied feedback—from targeted diagnostics to general writing suggestions. These strategies are typically categorized into: self-correction, correction with external feedback and multi-agent debate. More specifically, in self-correction methods such as Self-Refine \cite{madaan2023selfre}, Self-Verification \cite{gero2023selfver}, or Reflexion \cite{shinn2023reflexion} a single LLM both generates and refines its output iteratively until an acceptable quality of the output is obtained. Additionally, multiple external tools can be leveraged to provide improved feedback, thus leading to methods such as RAC \cite{li2024rac}, CRITIC \cite{gou2024critic}, FACTOOL \cite{chern2023factool}, REFINER \cite{paul2024refiner} and others \cite{pan2024survey}.

Our \textsc{FactCorrector} is a post-hoc factuality correction method, placing it in the same category as approaches like RAC and CRITIC. However, it distinguishes itself through its integration with \textsc{FactReasoner}, a recently introduced pipeline designed for long-form factuality assessment \cite{marinescu2025fr} which serves as the critic model. Furthermore, the refinement process in \textsc{FactCorrector} is guided by the probabilistic graphical model derived from \textsc{FactReasoner}.

\section{Preliminaries}
\label{sec-background}

We begin by providing background on long-form factuality assessment for LLMs.

\subsection{Long-Form Factuality Assessment}
Let $y$ be the long-form response generated by an LLM to a query $x$. We assume that $y$ can be decomposed into a set of $n$ \emph{atomic units} (or \emph{atoms}) that can be either true or false, denoted by $\cA_y = \{a_1, a_2, \ldots a_n\}$ \cite{factscore2023emnlp,song2024veriscore,wei2024longform}. An atomic unit $a_i \in \cA_y$ is defined as a short sentence conveying one piece of information (e.g., a claim or a fact). Furthermore, given an external knowledge source $\cK$\footnote{For example, $\cK$ could be Wikipedia, Google Search, or a collection of documents embedded into a vector database.}, we say that an atomic unit $a_i\in \cA_y$ is \emph{supported} by $\cK$ if there exists at least one piece of information in $\cK$ (e.g., a passage) called a \emph{context} that undebatably supports $a_i$. Otherwise, the atomic unit is \emph{not supported}.

The \emph{factual precision} denoted by $Pr(y)$ of the response $y$ with respect to a knowledge source $\cK$ is defined as: $Pr(y) = \frac{S(y)}{|\cA_y|}$, where $S(y) = \sum_{i=1}^{n} \mathbb{I}[a_i \textrm{ is supported by } \cK]$ is the number of supported atomic units. The \emph{factual recall} $R_K(y)$ up to the $K$-th supported atomic unit  is given by: $R_K(y) = \min(\frac{S(y)}{K}, 1)$. Finally, an $F_1$ measure for long-form factuality denoted by $F1@K$ can be defined as: $F_1@K(y) = \frac{2\cdot Pr(y) \cdot R_K(y)}{Pr(y) + R_K(y)}$ if $S(y) > 0$, and $0$ otherwise \cite{wei2024longform}.

\subsection{The \textsc{FactReasoner} Factuality Assessor}

\textsc{FactReasoner} \cite{marinescu2025fr} is a recent factuality assessor that, unlike previous prompt-based approaches \cite{factscore2023emnlp,song2024veriscore,wei2024longform}, leverages probabilistic reasoning to assess the factuality of the LLM generated response with respect to an external knowledge source $\cK$.

Specifically, the \textsc{FactReasoner} pipeline shown in Figure \ref{fig:fr-pipeline} (top) consists of four stages called \emph{Atomizer}, \emph{Reviser}, \emph{Retriever} and \emph{Evaluator}, respectively. The \emph{Atomizer} decomposes the input response $y$ into a set of $n$ atomic units $\cA_y$ by applying any of the decomposition strategies proposed recently \cite{factscore2023emnlp,bayat2025factbench}. The \emph{Reviser} post-processes the atoms such that the pronouns, unknown entities, or incomplete names are replaced with their corresponding named entities in the response \cite{wei2024longform}. Next, the \emph{Retriever} is responsible for querying the external knowledge source $\cK$ to retrieve the contexts relevant to the response's atoms \cite{song2024veriscore}. Finally, the \emph{Evaluator} constructs a graphical model representing a joint probability distribution over the atomic units in the response and their corresponding contexts in $\cK$. For each atom $a_i$, it then computes the posterior marginal probability $P(a_i)$ which quantifies the likelihood that $a_i$ is true (or supported) given the information available in $\cK$. Finally, the factuality metrics defined previously, such as $Pr(y)$, can be readily calculated using the supported atoms. Note that both the \emph{Atomizer} and \emph{Reviser} stages use LLMs for their specific tasks, as previously shown in \cite{marinescu2025fr}.

\paragraph{Graphical Model.} \textsc{FactReasoner}'s graphical model $\cG$ is defined by a tuple $\langle \bX, \bD, \bF \rangle$ where $\bX$ is a set of Boolean variables $A_i$ or $C_j$ associated with each atom $a_i \in \cA_y$ or context $c_j \in \cC_y$, respectively, where $\cC_y$ is the set of retrieved contexts. The domains $\bD$ are $\{\text{true}, \text{false}\}$ for all variables (e.g., we denote $a_i$ and $\neg a_i$ for $A_i=\text{true}$ and $A_i=\text{false}$, respectively) and $\bF$ is a set of factors. Each variable has a unary factor $f(\cdot)$ in $\bF$ encoding prior belief. Atoms have uniform priors: $f(a_i)=f(\neg a_i)=0.5$. Contexts from reliable sources have high prior (e.g., $f(c_j)=0.99$), while less reliable sources use smaller values. Binary factors $f(C_j,A_i)$ and $f(C_j,C_k)$ capture probabilistic logical relations between utterances. A relation model $p_\theta(\cdot|t,t')$ (either a specialized BERT model or an LLM) predicts the most likely relation from \{neutral, entail, contradict\} together with its probability. Therefore, each factor $f(C_j,A_i)$ or $f(C_j,C_k)$ represents a probabilistic encoding of an entailment relationship (i.e., context $C_j$ entails or supports atom $A_i$) or a contradiction relationship (i.e., context $C_j$ contradicts atom $A_i$), respectively. Note that the neutral relationships are ignored.

\begin{figure}[t]
    \centering
    \includegraphics[width=0.9\linewidth]{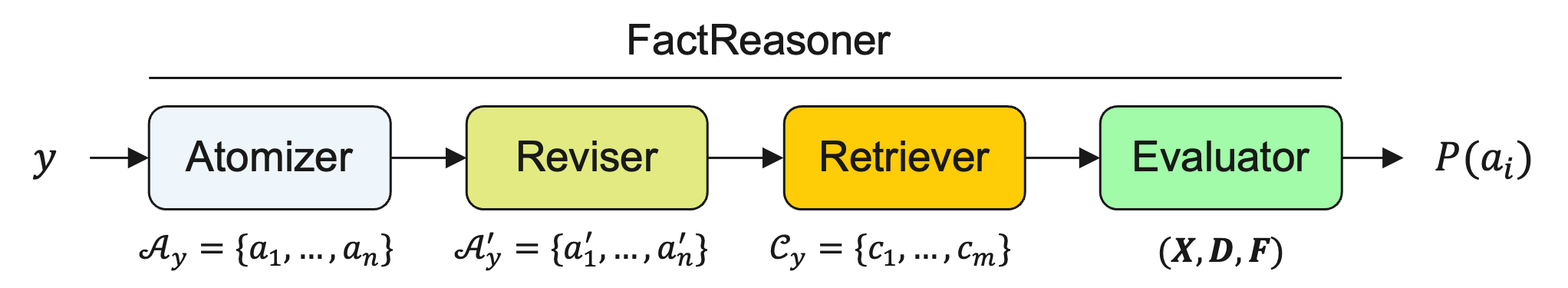}
    \includegraphics[width=0.8\linewidth]{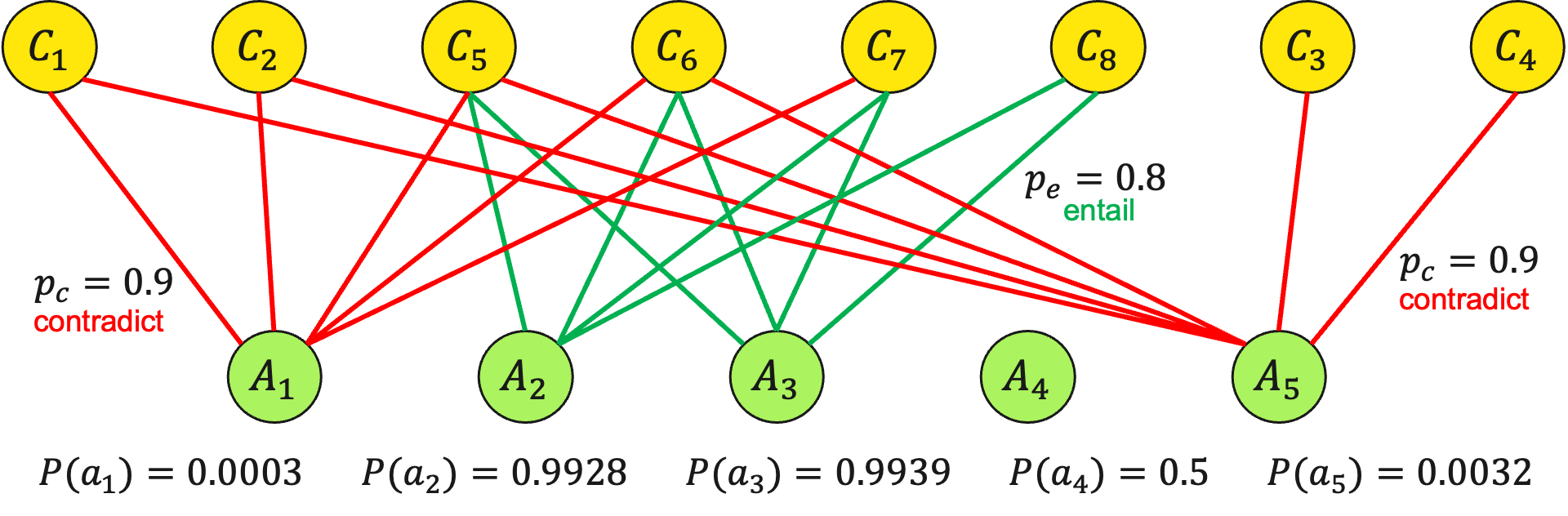}
    \caption{\textsc{FactReasoner} pipeline and an example graphical model with 5 atoms and 8 context variables.}
    \label{fig:fr-pipeline}
\end{figure}

\begin{example}
Figure \ref{fig:fr-pipeline} (bottom) shows an example graphical model with 5 atoms ($A_1$, ..., $A_5$), 8 contexts ($C_1$, ..., $C_8$) and 19 binary relations between atoms and contexts. Each edge in the graph is labeled by the probability of the corresponding entailment (green) or contradiction (red) relation.
\end{example}

\begin{figure*}[!ht]
    \centering
    \includegraphics[width=0.8\linewidth]{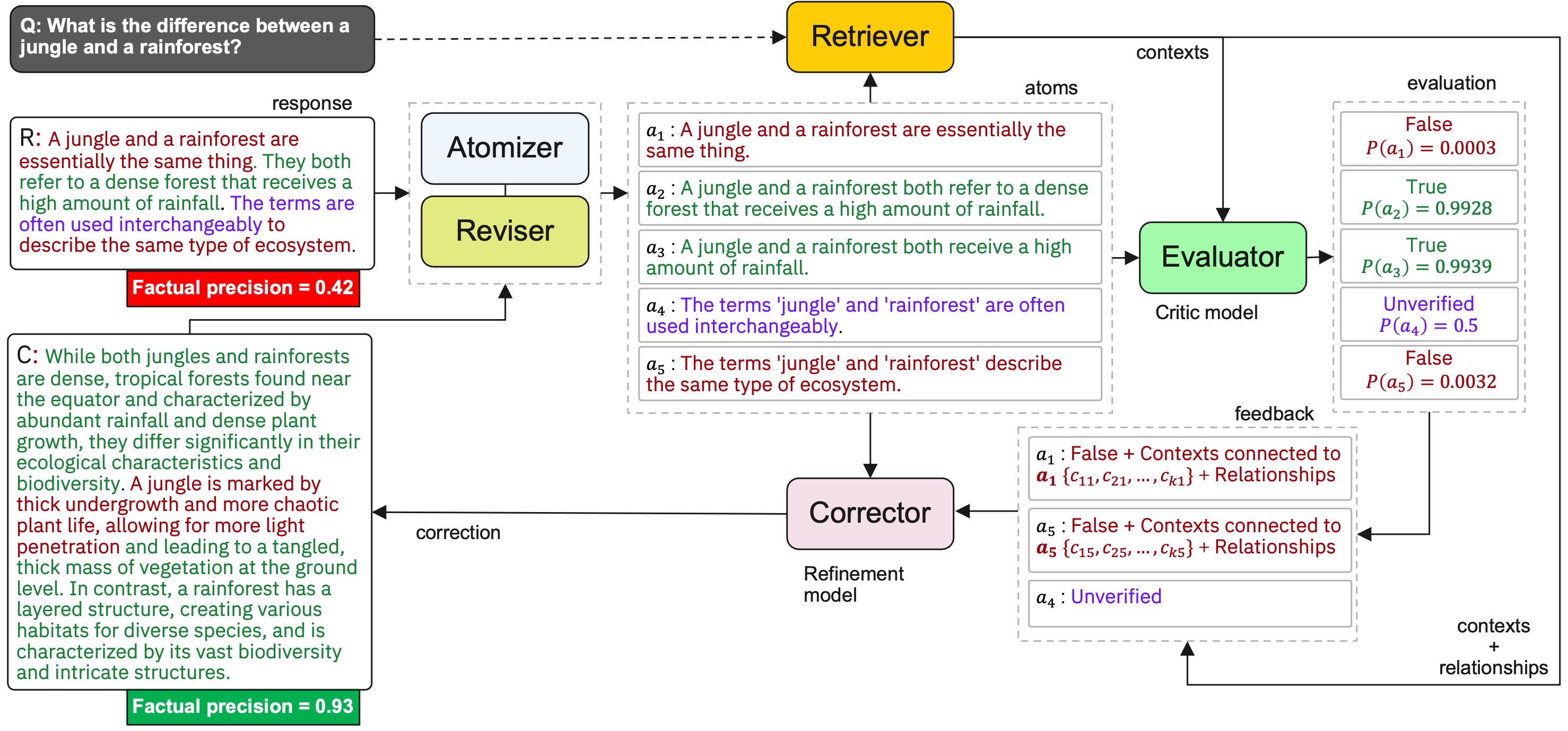}
    \caption{Illustration of the \textsc{FactCorrector} pipeline for long-form factuality correction of LLMs.}
    \label{fig:fc-pipeline}
\end{figure*}

\section{The \textsc{FactCorrector} Pipeline}
\label{sec-corrector}

\begin{algorithm}[t]

\begin{footnotesize}
\DontPrintSemicolon
  \KwIn{response $y$, refinement model $\cR$, threshold $\theta$}
  \KwOut{corrected response $y$}
  Evaluate factual precision $Pr(y)$ of $y$\;
  \While{$Pr(y) < \theta$}{
    Decompose $y$ into atoms $\cA_y = \{a_1, ..., a_n\}$\;
    Revise the atoms $\cA_y = \{a_1, ..., a_n\}$\;
    Retrieve contexts $\cC_y = \{c_1, ..., c_m\}$ from $\cK$\;
    Use $\cA_y$ and $\cC_y$ to build $\cG = \langle \bX,\bD,\bF \rangle$\;
    \For{atom $a_i\in \cA_i$}{
      Evaluate posterior probability $P(a_i)$ in $\cG$\;
      \lIf{$P(a_i) > 0.5$}{
        Label $a_i$ as True
      }\lElseIf{$P(a_i)<0.5$}{
        Label $a_i$ as False
      }\lElse{
        Label $a_i$ as Unverified
      }
    }
    $feedback \leftarrow \{a_i | a_i \textrm{ is False or Unverified }\} \cup \{c_j| \exists a_i\in \cG \textrm{ s.t. } c_j \textrm{ connects to } a_i \textrm{ in } \cG \textrm{ and } a_i \textrm{ is False}\}$\;

    $y' \leftarrow \cR(y, feedback)$\;
    Evaluate factual precision $Pr(y')$ of $y'$\;
    \If{$Pr(y') > Pr(y)$}{
        Let $y\leftarrow y'$ and $Pr(y) \leftarrow Pr(y')$
    }\lElse{
        \textbf{break}
    }
  }
  \Return{$y$}
\end{footnotesize}
\caption{\textsc{FactCorrector}}
\label{alg:fc}
\end{algorithm}

In this section, we present \textsc{FactCorrector}, a novel post-hoc method for correcting factual inaccuracies in long-form responses generated by LLMs. As a post-hoc correction strategy, \textsc{FactCorrector} employs a \emph{refinement model} $\mathcal{R}$ to revise an initial response $y$ to a user query $x$, guided by structured feedback from a \emph{critic model} $\mathcal{C}$ that evaluates the factual quality of the response. Specifically, the critic model in \textsc{FactCorrector} builds on the \textsc{FactReasoner} factuality assessor, which systematically identifies the incorrect parts within the response, while the refinement model can be any instruction-following LLM.

Algorithm \ref{alg:fc} describes the main steps of the proposed method which operate iteratively, starting from the initial response $y$. At each iteration, the critic decomposes the response into atomic units and retrieves relevant evidence from an external knowledge source $\cK$. These atoms and contexts are subsequently used to build a graphical model $\cG$ that captures entailment and contradiction relationships between atoms and retrieved contexts. Then, each atom $a_i$ is labeled as True, False, or Unverified if the posterior probability $P(a_i)$ is greater than, less than or equal to 0.5, respectively. The feedback includes all atoms labeled as False or Unverified, along with any contexts connected to them in $\cG$ and the corresponding relationship types, ensuring that corrections are grounded in explicit factual signals.

The refinement model then integrates this feedback with the original response in a structured prompt to produce a revised output. The process repeats until the corrected response meets a predefined factual precision threshold $\theta$ or there is no improvement.

\begin{example}
    Figure \ref{fig:fc-pipeline} illustrates the \textsc{FactCorrector} workflow on a simple question–response pair. In this example, the initial response has a factual precision of 0.42 and is decomposed into five atoms such that $a_2$ and $a_3$ are labeled as True, $a_1$ and $a_5$ as False, and $a_4$ as Unverified, respectively. The corresponding graphical model is shown in Figure \ref{fig:fr-pipeline} (bottom). The feedback includes the False atoms $\{a_1, a_5\}$ along with their contradicting and entailing contexts $\{c_1,c_2,...,c_7\}$. Atom $a_4$ is also flagged for removal. The refinement model (Corrector) uses this feedback to generate a corrected response that omits unverified claims and aims to correct the False atoms. The resulting output contains only one incorrect claim (highlighted in red) and achieves a factual precision of 0.93.
\end{example}

\textsc{FactCorrector} is closely related to the recent RAC method \cite{li2024rac}, but with a key difference: while RAC corrects each atom using only its retrieved documents, our approach uses the graphical model representing the atom-context relationships and aggregates all contexts connected to an incorrect atom into its feedback. This holistic view enables the refinement model to make more accurate corrections, reflected in the superior performance observed in our experiments.

\section{The \textsc{Veli5} Benchmark Dataset}
\label{sec-veli5}

In this section, we introduce the \textsc{Veli5} benchmark, a dataset specifically constructed to enable rigorous evaluation of long-form factuality correction methods and to facilitate the training of specialized models for correcting LLM generated responses.

The starting point for the VELI5 benchmark dataset is the ELI5-Category dataset\footnote{\href{https://huggingface.co/datasets/rexarski/eli5_category}{https://huggingface.co/datasets/rexarski/eli5\_category}}, a curated subset of ELI5~\cite{fan-etal-2019-eli5} that augments long-form explanatory question–answer threads scrapped from the \texttt{r/explainlikeimfive} reddit forum with explicit topical annotations. This resource contains questions in which users request intuitive explanations of complex topics, each assigned by community moderators to one of 12 high-level categories (11 topical domains plus a \textit{Repost} category) and paired with multiple candidate answers and their corresponding upvote scores.

For each question, we deterministically select the answer with the highest number of upvotes as the \emph{canonical response}. While these answers are generally high quality and well articulated, they are not guaranteed to be factually correct and may include inaccuracies, outdated claims, or speculative statements. Addressing this gap between explanatory quality and factual reliability is the primary motivation behind VELI5. We therefore apply \textsc{FactCorrector} to each canonical response.

To further diversify the dataset, we intentionally introduce factually incorrect responses to user questions. These synthetic responses are generated by prompting a reasonably strong LLM, such as the mixtral-8x22b-instruct model. The ratio of synthetic to human-authored responses is maintained at 50\%, ensuring a balanced mix of realistic and adversarial content. The final \textsc{Veli5} dataset comprises 17,522 instances, uniformly distributed across 12 categories. We further partition the dataset into training (14,017 samples), validation (1,752 samples), and test (1,753 samples) splits. Illustrative examples and additional details are provided in Appendix~\ref{sec-veli5-examples}.

\section{Experiments}
\label{sec-experiments}
We empirically evaluate the \textsc{FactCorrector} (FC) pipeline for long-form factuality correction and compare it against state-of-the-art approaches on several popular long-form factuality datasets.

\subsection{Baseline Correctors}
We consider two recent state-of-the-art post-hoc correction methods with feedback: CRITIC \cite{gou2024critic} and RAC \cite{li2024rac}. CRITIC is a prompt-based iterative approach that leverages LLMs and retrieved knowledge. At each step, the model generates a query to retrieve relevant information, revises its output based on revision history, and selects the most plausible answer. RAC, in contrast, decomposes the generated response into atomic facts, retrieves evidence from trusted sources (e.g., Google), verifies each fact, and corrects inaccuracies using reliable content. In addition, we also evaluate two prompting strategies: LLM1, which performs corrections using only the model’s internal knowledge, and LLM2, which uses contexts retrieved for the question only and ignores its internal knowledge (see Appendix \ref{sec-prompts}).

We instantiated the competing correctors with open-source LLMs from the IBM Granite~\cite{granite-paper}, Meta LLaMA~\cite{touvron2023llama}, MistralAI Mixtral~\cite{mixtral-paper} and OpenAI~\cite{gpt-paper} families, spanning a broad range of architectures and scales. All models run remotely on A100 80GB GPUs and are accessed via \texttt{litellm} APIs (1,500 prompts per sec).

\subsection{Datasets}
We use three datasets in our experiments: a reduced version of \textsc{Veli5} (200 instances uniformly sampled from the test split), Biographies (\textsc{Bio})~\cite{factscore2023emnlp}, and AskHistorians (\textsc{AskHist})~\cite{xu2023}.
The \textsc{Bio} dataset includes 183 biographical passages generated by ChatGPT for entities with Wikipedia pages. \textsc{AskHist} contains 200 questions from the r/AskHistorians reddit forum paired with long-form answers produced by the llama-3.3-70b-instruct model.

We also use the \textsc{Conflicts} dataset from \cite{marinescu2025fr}, which comprises 100 atomic claims (or responses) sampled from ConflictBank~\cite{su2024conflictbank}. Each claim, originally from Wikidata, is assumed true and is associated with both supporting and conflicting contexts. Since all claims are correct, they require no corrections, offering a controlled setting for our evaluation.

\subsection{Measures of Performance}
For each dataset $\mathcal{D}$ and each competing corrector, we computed three factuality metrics: precision ($Pr$), recall at $K$ ($R@K$), and $F_1@K$, averaged over all prompts in $\mathcal{D}$, where $K$ is set to the median number of atoms. These metrics are evaluated using the \textsc{FactReasoner} (FR) assessor with Google search results as external knowledge source \cite{marinescu2025fr}. In addition, we computed two complementary metrics: verifiability ($V$) and comprehensiveness ($C$). Verifiability $V$ is defined as the number of atoms in the response (or correction) that can be verified—i.e., atoms connected to either supporting or contradicting evidence in FR’s graph. Comprehensiveness $C$ quantifies coverage and is given by $C = \frac{|\cA_{in}|}{|\cA_{in}| + |\cA_{out}|}$, where $\cA_{in}$ denotes the atoms covered by the response (or correction), and $\cA_{out}$ denotes atoms that are \emph{uncovered} or missing~\cite{dejl2025compr}.

For each factuality metric $S$ (e.g., precision), we report its \emph{relative gain}, denoted as $G(S)$ and defined by:
$G(S) = \frac{2 \cdot (S_c - S_r)}{S_c + S_r}$, where $S_r$ and $S_c$ are the metrics corresponding to the original response and the correction, respectively. A positive $G(S)$ indicates that the correction outperforms the response, while a negative value means that the correction performs worse. By construction, $G(S)$ ranges from $-2$ to $2$ and remains well defined even when either $S_r$ or $S_c$ equals zero.

\subsection{Results for Post-hoc Correction}
We next present the results obtained on our datasets using a range of LLMs. For consistency, the same LLM employed by \textsc{FactReasoner} to evaluate the metrics reported in the tables was also used to instantiate the components of \textsc{FactCorrector}.

\begin{table}[t!]
    \centering
    \resizebox{0.9\linewidth}{!}{
    \begin{tabular}{l|c|c|c|c}
    corrector & ROUGE $\uparrow$ & BLEU $\uparrow$ & BLEURT $\uparrow$ & JUDGE $\uparrow$ \\
    \toprule
    \multicolumn{5}{c}{\texttt{mixtral-8x22b-instruct}}\\
    \midrule
    CRITIC &  0.14$\pm$0.07   & 0.02$\pm$0.03   & -0.69$\pm$0.25   & 0.41  \\
    RAC    &  \underline{0.54}$\pm$0.23   & \underline{0.23}$\pm$0.27   &  \underline{0.29}$\pm$0.43   & {\bf 0.87}  \\
    LLM1   &  0.15$\pm$0.05    & 0.02$\pm$0.02   & -0.63$\pm$0.21   & 0.30  \\
    LLM2   &  0.20$\pm$0.10    & 0.05$\pm$0.05   & -0.48$\pm$0.22   & 0.10  \\
    \midrule
    FC (ours) & {\bf 0.89$\pm$0.26} & {\bf 0.87$\pm$0.32} & {\bf 0.73$\pm$0.47} & {\bf 0.87} \\
    \midrule
    \multicolumn{5}{c}{\texttt{llama-3.3-70b-instruct}}\\
    \midrule
    CRITIC &  0.32$\pm$0.27   &  0.15$\pm$0.28  &  -0.36$\pm$0.56   & 0.31  \\
    RAC    &  {\bf 0.87$\pm$0.20}   &  {\bf 0.77$\pm$0.32}  &  {\bf 0.65$\pm$0.43}   & {\bf 0.77}  \\
    LLM1   &  0.19$\pm$0.11   &  0.03$\pm$0.05  & -0.62$\pm$0.33    & 0.17  \\
    LLM2   &  0.21$\pm$0.10   &  0.04$\pm$0.04  & -0.46$\pm$0.22    & 0.06  \\
    \midrule
    FC (ours) & \underline{0.73}$\pm$0.33 & \underline{0.62}$\pm$0.45   & \underline{0.46}$\pm$0.56   & \underline{0.60} \\
    \midrule
    \multicolumn{5}{c}{\texttt{granite-4.0-h-small}}\\
    \midrule
    CRITIC &  0.15$\pm$0.11   & 0.03$\pm$0.10   & -0.69$\pm$0.26   & 0.20  \\
    RAC    &  \underline{0.74}$\pm$0.23   & \underline{0.48}$\pm$0.37   & \underline{0.49}$\pm$0.46   &  \underline{0.73} \\
    LLM1   &  0.15$\pm$0.07   & 0.02$\pm$0.02   & -0.68$\pm$0.23   & 0.17  \\
    LLM2   &  0.42$\pm$0.24   & 0.15$\pm$0.17   & -0.10$\pm$0.44   & 0.13  \\
    \midrule
    FC (ours) & {\bf 0.83$\pm$0.30} & {\bf 0.76$\pm$0.39}   & {\bf 0.58$\pm$0.57}   & {\bf 0.78} \\
    \midrule

    \end{tabular}
    }
    \caption{Results obtained on the \textsc{Conflicts} dataset.}
    \label{tab:conflicts}
\end{table}

\paragraph{\textsc{Conflicts}.} Table \ref{tab:conflicts} summarizes the results obtained on our controlled experiment with the \textsc{Conflicts} dataset. In this case, we assess the similarity between correction and response using the ROUGE, BLEU, and BLEURT metrics~\cite{lin2004rouge,papineni2002bleu,sellam2020bleurt} because the response is known to be true and doesn't need correction. We also report JUDGE, which counts the number of instances where an LLM-as-a-Judge model (in our case DeepSeek-3.2), prompted appropriately, infers that the correction is equivalent to the original response. The table shows the mean and standard deviation for all metrics. The results indicate that \textsc{FactCorrector} achieves superior performance with the Mixtral and Granite models, while ranking second when using LLaMA, where RAC attains the best scores. In contrast, the LLM1 and LLM2 baselines perform rather poorly almost always producing corrections significantly different from the response, while CRITIC often struggles to follow its instructions also leading to poor results.

\begin{table}[t!]
    \centering
    \resizebox{0.7\linewidth}{!}{
    \begin{tabular}{l|c|c|c|c|c}
    corrector &  Pr $\uparrow$   & R@K $\uparrow$ & F1@K $\uparrow$ & V $\uparrow$ & C $\uparrow$ \\
\toprule
\multicolumn{6}{c}{\texttt{mixtral-8x22b-instruct}} \\
\midrule
CRITIC  & 0.26 & 0.08  & 0.19  & 0.09   &  0.06 \\
RAC     & 0.24 & 0.08  & 0.18  & 0.08   &  0.06 \\
LLM1    & \underline{0.32} & {\bf 0.10}  & \underline{0.23}  & 0.12   &  {\bf 0.07} \\
LLM2    & \underline{0.32} & {\bf 0.10}  & \underline{0.23}  & \underline{0.13}   &  {\bf 0.07} \\
\midrule
FC (ours)  & {\bf 0.34} & \underline{0.09}  & {\bf 0.24}  & {\bf 0.14}   &  {\bf 0.07} \\
\midrule
\multicolumn{6}{c}{\texttt{llama-3.3-70b-instruct}} \\
\midrule
CRITIC  & 0.19 & 0.06  & 0.13  & 0.03   &  0.05 \\
RAC     & 0.13 & 0.02  & 0.09  & 0.02   &  0.05 \\
LLM1    & 0.25 & \underline{0.06}  & {\bf 0.18}  & {\bf 0.05}   &  {\bf 0.08} \\
LLM2    & \underline{0.26} & {\bf 0.07}  & {\bf 0.18}  & {\bf 0.05}   &  {\bf 0.08} \\
\midrule
FC (ours)  & {\bf 0.27} & {\bf 0.07}  & {\bf 0.18}  & {\bf 0.05}   &  {\bf 0.08} \\
\midrule
\multicolumn{6}{c}{ \texttt{granite-4.0-h-small}} \\
\midrule
CRITIC  & 0.17 & 0.04  & 0.12  & 0.05   &  0.06 \\
RAC     & 0.20 & 0.09  & 0.16  & 0.06   &  \underline{0.07} \\
LLM1    & {\bf 0.31} & {\bf 0.11}  & {\bf 0.23}  & {\bf 0.11}   &  {\bf 0.08} \\
LLM2    & 0.20 & \underline{0.10}  & 0.16 & \underline{0.09} &  0.05 \\
\midrule
FC (ours)  & \underline{0.29} & \underline{0.10}  & \underline{0.21}  & {\bf 0.11}   &  \underline{0.07} \\
\midrule
\multicolumn{6}{c}{\texttt{gpt-oss-120b}} \\
\midrule
CRITIC  & -0.04 & 0.01  & -0.02  & -0.01    &  -0.03 \\
RAC     & \underline{0.34}  & 0.15  & 0.27   & \underline{0.06}     &  \underline{0.13} \\
LLM1    & 0.21 & 0.16   & 0.19   & -0.01    &  0.13 \\
LLM2    & 0.33 & {\bf 0.20}   & \underline{0.28}   & 0.05     &  \underline{0.13} \\
\midrule
FC (ours)  & {\bf 0.36} & \underline{0.19}  & {\bf 0.30}  & {\bf 0.07}   &  {\bf 0.14} \\
\midrule
    \end{tabular}
    }
    \caption{Mean relative gains for the factuality metrics obtained on the \texttt{VELI5} dataset.}
    \label{tab:veli5-gains}
\end{table}

\begin{figure}[t]
    \centering
    \includegraphics[width=0.9\linewidth]{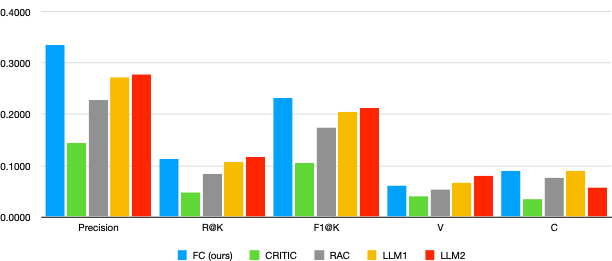}
    \caption{Mean relative gains for factuality metrics across models on the \textsc{Veli5} dataset.}
    \label{fig:veli5-gains-macro}
\end{figure}

\begin{table}[t]
    \centering
    \resizebox{0.9\linewidth}{!}{
    \begin{tabular}{l|cc|cc|cc|cc}
    corrector &  \multicolumn{2}{c|}{llama-3.3-70b} & \multicolumn{2}{c|}{mixtral-8x22b} & \multicolumn{2}{c|}{granite-4.0-small} & \multicolumn{2}{c}{gpt-oss-120b} \\
    & before & after & before & after & before & after & before & after \\
\toprule
CRITIC  & 0.78 & 0.90  & 0.70  & 0.88   &  0.74  & 0.86 & 0.62 & 0.58 \\
RAC     & 0.78 & 0.91  & 0.70  & 0.86   &  0.74  & 0.88 & 0.62 & {\bf 0.85} \\
LLM1    & 0.78 & 0.95  & 0.70  & \underline{0.91}   &  0.74  & {\bf 0.94} & 0.62 & 0.72 \\
LLM2    & 0.78 & \underline{0.96}  & 0.70  & \underline{0.91}   &  0.74  & 0.86 & 0.62 & 0.79 \\
\midrule
FC (ours)  & 0.78 & {\bf 0.97}  & 0.70  & {\bf 0.93}   &  0.74 & \underline{0.93} & 0.62 & \underline{0.84} \\
\midrule
\end{tabular}
}
\caption{Mean factual precision before and after correction on the \textsc{VELI5} dataset.}
\label{tab:veli5-precision}
\end{table}

\paragraph{\textsc{Veli5}.} In Table \ref{tab:veli5-gains} we report the mean relative gains obtained for the factuality metrics on the \textsc{Veli5} dataset. The results demonstrate that \textsc{FactCorrector} consistently delivers strong and reliable improvements in factuality across all evaluated metrics and models. For example, when considering relative gains on precision, FC outperforms all baselines, achieving macro-averaged gains of 0.315 — substantially higher than the next-best corrector (see also Figure \ref{fig:veli5-gains-macro}).  Importantly, FC is often best or tied-best in many metric–model combinations, showing strong generalization across diverse LLM architectures. LLM1 and LLM2 show strong performance in this case, likely due to the provenance of the VELI5 questions which were originally derived from a public Reddit forum. Therefore, it is plausible that similar content was included in these models’ pre-training corpora which may confer a significant advantage on this dataset.

Table \ref{tab:veli5-precision} reinforces these findings by examining factual precision before and after correction. FC achieves the highest post-correction precision on two out of four models (Llama, Mixtral) and ranks second on Granite and GPT-OSS, where it still delivers large relative gains. On average, FC improves precision by +0.21, outperforming all baselines, including LLM1 (+0.17) and RAC (+0.16). While RAC slightly surpasses FC on GPT-OSS in final precision, FC dominates in overall gains, suggesting a more robust correction strategy. In contrast, CRITIC exhibits inconsistent behavior, even reducing precision for GPT-OSS.

\paragraph{\textsc{Bio}.} Table \ref{tab:bio-gains} presents the mean relative gains on the \textsc{Bio} dataset. FC exhibits robust performance across models, achieving its highest gains with Mixtral and Granite, and ranking second for Llama. In contrast, LLM1 performs poorly, particularly on Granite and GPT-OSS. Both RAC and LLM2 remain highly competitive on this dataset, while CRITIC, as observed previously, struggles to follow instructions, leading to inconsistent results.

\begin{table}[t!]
    \centering
    \resizebox{0.7\linewidth}{!}{
    \begin{tabular}{l|c|c|c|c|c}
    corrector &  Pr $\uparrow$   & R@K $\uparrow$ & F1@K $\uparrow$ & V $\uparrow$ & C $\uparrow$ \\
\toprule
\multicolumn{6}{c}{\texttt{mixtral-8x22b-instruct}} \\
\midrule
CRITIC  & 0.27 & 0.22 & 0.25  & 0.05  & 0.05  \\
RAC     & 0.39 & 0.29 & 0.34  & 0.05  & 0.13  \\
LLM1    & 0.27 & 0.25 & 0.27  & 0.02  & \underline{0.07}  \\
LLM2    & \underline{0.42} & \underline{0.30} & \underline{0.37}  & \underline{0.11}  & \underline{0.07}  \\
\midrule
FC (ours)  & {\bf 0.46} & {\bf 0.33}  & {\bf 0.41}  & {\bf 0.12}   &  {\bf 0.11} \\
\midrule
\multicolumn{6}{c}{\texttt{llama-3.3-70b-instruct}} \\
\midrule
CRITIC  & 0.37 & 0.18 & 0.29  & 0.08  & 0.13  \\
RAC     & 0.42 & 0.21 & 0.33  & 0.12  & 0.16  \\
LLM1    & 0.23 & 0.14 & 0.19  & 0.07  & 0.11  \\
LLM2    & {\bf 0.45} & \underline{0.21} & {\bf 0.36}  & {\bf 0.12}  & {\bf 0.18}  \\
\midrule
FC (ours)  & \underline{0.44} & {\bf 0.22}  & \underline{0.35}  & {\bf 0.12}   &  \underline{0.17} \\
\midrule
\multicolumn{6}{c}{ \texttt{granite-4.0-h-small}} \\
\midrule
CRITIC  & 0.12 & 0.02 & 0.08  & 0.04  & 0.02  \\
RAC     & \underline{0.22} & 0.08 & 0.16 & 0.10  & 0.04  \\
LLM1    & -0.03 & -0.01 & -0.02 & 0.02 & 0.00  \\
LLM2    & 0.19 & {\bf 0.13} & \underline{0.16} & {\bf 0.13}  & {\bf 0.03}  \\
\midrule
FC (ours)  & {\bf 0.26} & \underline{0.10}  & {\bf 0.20}  & \underline{0.14}   & {\bf 0.03} \\
\midrule
\multicolumn{6}{c}{\texttt{gpt-oss-120b}} \\
\midrule
CRITIC  & 0.00  & 0.00 & 0.00  & 0.00  & 0.01  \\
RAC     & {\bf 0.51}  & 0.28 & {\bf 0.42}  & {\bf 0.11}  & {\bf 0.28}  \\
LLM1    & -0.28 & -0.10 & -0.22  & -0.24  & 0.06  \\
LLM2    & \underline{0.48} & {\bf 0.31} & \underline{0.41}  & \underline{0.10}  & \underline{0.27}  \\
\midrule
FC (ours)  & 0.40 & \underline{0.29}  & 0.36  & \underline{0.10}  &  0.24 \\
\midrule
    \end{tabular}
    }
    \caption{Mean relative gains for the factuality metrics obtained on the \textsc{Bio} dataset.}
    \label{tab:bio-gains}
\end{table}

\begin{table}[t!]
    \centering
    \resizebox{0.7\linewidth}{!}{
    \begin{tabular}{l|c|c|c|c|c}
    corrector &  Pr $\uparrow$   & R@K $\uparrow$ & F1@K $\uparrow$ & V $\uparrow$ & C $\uparrow$ \\
\toprule
\multicolumn{6}{c}{\texttt{mixtral-8x22b-instruct}} \\
\midrule
CRITIC  & -0.12 & -0.04 & -0.09 & -0.06 & -0.01 \\
RAC     & \underline{-0.01} & -0.01 & -0.01 & -0.01 & 0.00 \\
LLM1    & -0.02 & 0.00 & -0.01 & 0.00 & 0.00 \\
LLM2    & -0.03 & 0.00 & -0.02 & 0.00 & 0.00 \\
\midrule
FC (ours)  & {\bf 0.02} & {\bf 0.00}  & {\bf 0.00} & {\bf 0.02} & {\bf 0.00} \\
\midrule
\multicolumn{6}{c}{\texttt{llama-3.3-70b-instruct}} \\
\midrule
CRITIC  & -0.03 & 0.00  & -0.01 & -0.01 & -0.01 \\
RAC     & \underline{0.03} & 0.01 & 0.02 & 0.02 & 0.00 \\
LLM1    & 0.00 & 0.00 & 0.00 & 0.00 & 0.00 \\
LLM2    & 0.01 & 0.00 & 0.01 & 0.00 & 0.00 \\
\midrule
FC (ours)  & {\bf 0.04} & {\bf 0.01}  & {\bf 0.02} & {\bf 0.01} & {\bf 0.01} \\
\midrule
\multicolumn{6}{c}{ \texttt{granite-4.0-h-small}} \\
\midrule
CRITIC  & -0.13 & -0.09 & -0.11 & -0.05 & -0.04 \\
RAC     & \underline{-0.06} & -0.01 & -0.04 & -0.01 & -0.02 \\
LLM1    & -0.08 & -0.03 & -0.05 & -0.03 & -0.02 \\
LLM2    & -0.07 & -0.01 & -0.04 & -0.03 & -0.01 \\
\midrule
FC (ours)  & {\bf 0.00} & {\bf 0.00}  & {\bf 0.00} & {\bf 0.01} & {\bf 0.00} \\
\midrule
\multicolumn{6}{c}{\texttt{gpt-oss-120b}} \\
\midrule
CRITIC  & -0.03 & -0.01 & -0.02 & 0.01 & -0.01 \\
RAC     & {\bf 0.12} & {\bf 0.03} & {\bf 0.08} & {\bf 0.06} & {\bf 0.02} \\
LLM1    & -0.49 & -0.26 & -0.41 & -0.19 & -0.08 \\
LLM2    & -0.03 & -0.01 & -0.03 & 0.03 & -0.04 \\
\midrule
FC (ours)  & \underline{0.00} & 0.00  & 0.00 & 0.01 & 0.00 \\
\midrule
    \end{tabular}
    }
    \caption{Mean relative gains for the factuality metrics obtained on the \textsc{AskHist} dataset.}
    \label{tab:askhist-gains}
\end{table}

\paragraph{\textsc{AskHist}.} Table \ref{tab:askhist-gains} shows the mean relative gains on the \textsc{AskHist} dataset. Unlike previous cases, the gains here are smaller. This is primarily because the factual precision of the original responses is already high (typically exceeding 89\% on average) leaving limited room for improvement through correction. Interestingly, all other baselines exhibit negative gains, indicating that their corrections are less factual than the original responses. In contrast, FC maintains robust performance across all models, almost always producing corrections that are at least as factual as the original responses and, in many cases, slightly better.

\begin{table}[t!]
    \centering
    \resizebox{0.7\linewidth}{!}{
    \begin{tabular}{l|c|c|c|c|c}
    corrector &  Pr $\uparrow$   & R@K $\uparrow$ & F1@K $\uparrow$ & V $\uparrow$ & C $\uparrow$ \\
\toprule
\multicolumn{6}{c}{\texttt{mixtral-8x22b-instruct}} \\
\midrule
FC (ours)  & {\bf 0.34} & 0.09  & {\bf 0.24}  & {\bf 0.14}   &  0.07 \\
SFT (ours) &     0.19 & {\bf 0.10}  & 0.16  & 0.05   &  {\bf 0.08} \\
\midrule
\multicolumn{6}{c}{\texttt{llama-3.3-70b-instruct}} \\
\midrule
FC (ours)  & {\bf 0.27} & 0.07  & 0.18  & 0.05   &  {\bf 0.08} \\
SFT (ours) & 0.25 & {\bf 0.11}  & 0.19  & {\bf 0.10}   &  {\bf 0.08} \\
\midrule
\multicolumn{6}{c}{ \texttt{granite-4.0-h-small}} \\
\midrule
FC (ours)  & {\bf 0.29} & 0.10  & {\bf 0.21}  & {\bf 0.11}   &  {\bf 0.07} \\
SFT (ours) & 0.22 & {\bf 0.11}  & 0.18  & 0.09   &  {\bf 0.07} \\
\midrule
\multicolumn{6}{c}{\texttt{gpt-oss-120b}} \\
\midrule
FC (ours)  & {\bf 0.36} & {\bf 0.19}  & {\bf 0.30}  & {\bf 0.07}   &  {\bf 0.14} \\
SFT (ours) & 0.08 & 0.08  & 0.08  & {\bf 0.07}   &  0.02\\
\midrule
    \end{tabular}
    }
    \caption{Mean relative gains for the factuality metrics on the \texttt{VELI5} dataset using the SFT corrector.}
    \label{tab:sft-veli5}
\end{table}

\subsection{Results for SFT Correction}
We used the larger \textsc{Veli5} dataset to train a LoRA adapter on the Granite-Guardian-5B model for generating corrections\footnote{LoRA adapter available at \href{https://ibm.biz/granite-guardian-3-2-5b-lora-factuality-correction}{https://ibm.biz/granite-guardian-3-2-5b-lora-factuality-correction}.}. Table \ref{tab:sft-veli5} reports the mean relative gains on the \textsc{Veli5} dataset when using the \textsc{FactCorrector} pipeline versus the LoRA-based correction approach (denoted as SFT). Overall, the SFT method produces substantially better corrections than the original responses, which is expected given that the evaluation instances originate from the same distribution as the training data. In contrast, on the out-of-distribution \textsc{Bio} and \textsc{AskHist} datasets the gains achieved by SFT are smaller than before; however, most gains remain non-negative, indicating that the SFT approach generalizes reasonably well (see Appendix \ref{sec-experiments-all}).

\subsection{Human Evaluation of \textsc{Veli5}}

We conducted a user study to evaluate the quality of corrections in the \textsc{Veli5} dataset (see Appendix \ref{sec-human-eval}). Table \ref{tab:atom-eval-all} summarizes results for 30 filtered tasks, each averaging 2.30 incorrect and 4.40 correct atoms. At the task level, original responses were highly relevant to the user question (90\% fully, 10\% partially), while corrections achieved even greater relevance (96.7\% fully). For incorrect atoms, 67.4\% were successfully corrected by \textsc{FactCorrector}, with 26.4\% undercorrected and only minor proportions overcorrected (3.3\%) or invalid (2.9\%). Correct atoms were preserved in 60.9\% of cases, with 27.7\% not preserved and small fractions partially preserved or invalid. Relevance analysis showed both incorrect and correct atoms were predominantly relevant (75.0\% and 72.4\%), underscoring the effectiveness of the correction process.

\section{Conclusion}
\label{sec-conclusion}

The contributions of this paper are threefold. (1) We introduce \textsc{FactCorrector}, a novel post-hoc method for improving long-form factuality by leveraging feedback on incorrect response segments and retrieved contradicting evidence to generate refined answers. (2) We develop \textsc{Veli5}, a benchmark dataset of curated question–answer pairs explicitly verified and corrected using \textsc{FactCorrector}, enabling rigorous evaluation. (3) We conduct extensive empirical studies across a wide range of open-source LLMs and datasets, demonstrating that \textsc{FactCorrector} consistently achieves the most reliable factuality improvements compared with strong state-of-the-art baselines.

\section*{Limitations}
We acknowledge further limitations of the proposed \textsc{FactCorrector} framework.

The Atomizer and Reviser components are sensitive to the quality of the prompt and few shot examples used as well as the LLM employed to perform the atomic unit decomposition and decontextualization of the response (and correction). In our work we only considered open-source models from the Mixtral, Llama, Granite and GPT-OSS families. Furthermore, decomposing a given response can be done at different granularities such as sentence level, paragraph level or the entire response level. Our implementation is limited to decomposing the response in one shot.

The quality of the contexts retrieved by the Retriever component for each atomic unit depends on both the implementation details of the component and the formulation of the query string it receives. In our approach, we leverage an LLM to generate a well-structured Google search query for each atomic unit, which is then used to retrieve the top $k$ search results via the Serper API (serper.dev). Each search result typically includes a short snippet and a hyperlink; to enrich the retrieved context, we additionally extract up to 4,000 characters of text from the linked page.

The Evaluator component relies on the logical relationships between atoms and contexts. It also depends on the quality of the prompt and the underlying LLM. As before, we only used open-source models such as mixtral-8x22b-instruct, llama-3.3-70b-instruct, granite-4.0-h-small, and gpt-oss-120b with a fairly straightforward prompt. It is possible to craft better prompts that could lead to a better extraction of the relationships. Fine-tuning is another option to obtain a stronger relation model.

The Corrector component (i.e., refinement model) is also sensitive to the quality of the prompt used. The version presented in Table \ref{tab:prompt-fc} in Appendix \ref{sec-prompts} is the current version that works farily well across different models. Clearly, improved prompts could lead to much better corrections and this is currently one direction of future research.

Finally, since the \textsc{FactCorrector} pipeline is tightly integrated with the \textsc{FactReasoner} factuality assessor \cite{marinescu2025fr}, it inherits the computational overhead of the latter. Specifically, the Evaluator component requires $O(n\cdot m)$ LLM calls to extract the logical relationships between atoms and contexts, where $n$ is the number of atomic units in the reponse (or correction), $m$ is the total number of non-duplicated contexts retrieved for the atoms. In contrast, the SFT approach for correction requires only a single LLM call per correction.

\section*{Ethical Statement}
We recognize the positive and negative societal impacts of LLMs in general, including potential misuse of our work around factuality correction of LLM generated output. We note that the datasets considered are public and peer reviewed, there are no human subjects involved, and as far as we know, there are no obvious harmful consequences from our work. All creators and original owners of assets have been properly credited and licenses and terms of use have been respected. We have not conducted crowd-sourcing experiments or research with human subjects.

\bibliography{ref}

\newpage

\appendix

\section{Background on Graphical Models}
\label{apx:pgm}
Graphical models such as Bayesian or Markov networks provide a powerful framework for reasoning about conditional dependency structures over many variables \cite{pearl88,koller2009probabilistic}.

A \emph{graphical model} is a tuple $\cM = \langle \bX,\bD,\bF \rangle$, where $\bX = \{X_1, \ldots, X_n\}$ is a set of variables, $\bD = \{D_1, \ldots, D_n\}$ is the set of their finite domains of values and $\bF =\{f_1, \ldots, f_m\}$ is a set of discrete positive real-valued functions. Each function $f_i$ (also called \emph{factor}) is defined on a subset of variables $\bS_i \subseteq \bX$ called its \emph{scope} and denoted by $vars(f_i)$. The model $\cM$ defines a factorized probability distribution on $\bX$:

\begin{align}\label{eq:pgm}
P(\bx) = \frac{1}{Z}\prod_{j=1}^{m} f_j(\bx) ~\textrm{s.t.}~ & Z = \sum_{\bx \in \Omega(\bX)} \prod_{j=1}^{m} f_j(\bx)
\end{align}

\noindent where the normalization constant $Z$ is known as the \emph{partition function} and $\Omega(\bX)$ denotes the Cartesian product of the variables domains.

The function scopes of a model $\cM$ define a \emph{primal graph} whose vertices are the variables and its edges connect any two variables that appear in the scope of the same function.

A common inference task over graphical models is to compute the posterior marginal distributions over all variables. Namely, for each variable $X_i \in \bX$ and domain value $x_i \in D_i$, compute:

\begin{align}\label{eq:marginal}
P(x_i) &= \sum_{\bx \in \Omega(\bX)} \delta_{x_i}(\bx)\cdot P(\bx)
\end{align}

\noindent where $\delta_{x_i}(\bx)$ is $1$ if $X_i$ is assigned $x_i$ in $\bx$ and $0$ otherwise \cite{koller2009probabilistic}.

\begin{figure}
    \centering
    \includegraphics[width=1.0\linewidth]{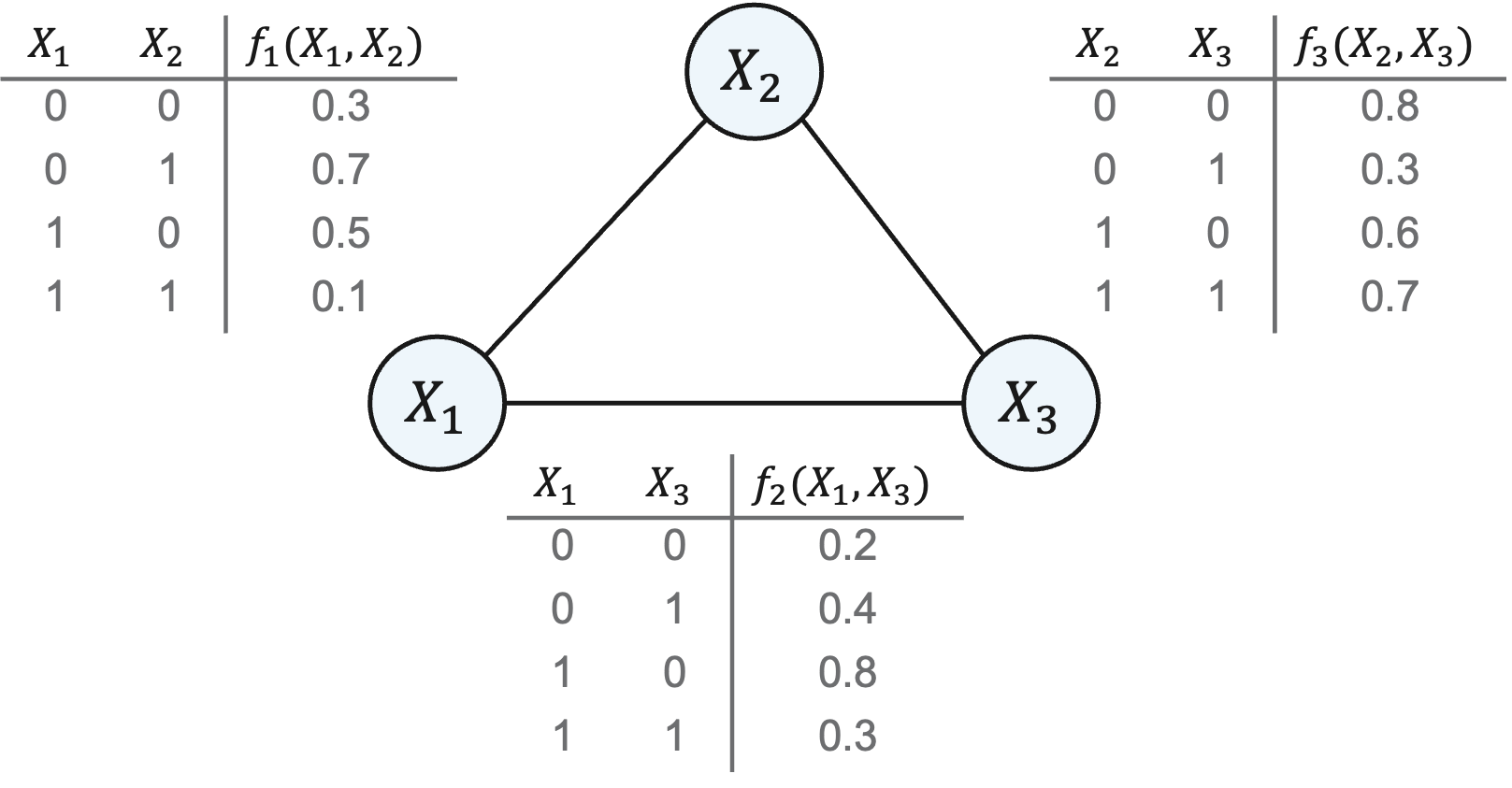}
    \caption{A graphical model with three bi-valued variables $X_1$, $X_2$ and $X_3$, and three binary functions.}
    \label{fig:pgm}
\end{figure}

\begin{example}
 Figure \ref{fig:pgm} shows a graphical model with 3 bi-valued variables $X_1$, $X_2$ and $X_3$ and 3 binary functions $f_1(X_1,X_2)$, $f_2(X_1,X_3)$ and $f_3(X_2,X_3)$. The joint probability distribution is given by $P(X_1,X_2,X_3)=\frac{1}{Z} \cdot f_1(X_1,X_2)\cdot f_2(X_1,X_3)\cdot f_3(X_2,X_3)$. In this case, the posterior marginal distribution of $X_1$ is: $P(X_1=0) = 0.46$ and $P(X_1=1) = 0.54$, respectively.
\end{example}

Equation \ref{eq:marginal} can be solved using any probabilistic inference algorithm for graphical models, such as variable elimination \cite{dechter-book}, belief propagation \cite{pearl88}, or variational inference \cite{liu2011}. In our implementation, we employed the Weighted Mini-Buckets (WMB) algorithm \cite{liu2011}. WMB is parameterized by an i-bound, which controls the trade-off between computational complexity and inference accuracy. For our experiments, we selected an i-bound of 6, which enabled us to solve all inference problems efficiently. Notably, WMB proved highly effective in practice, solving each inference instance in under 0.05 seconds across our benchmark datasets.

\section{Background on Factuality Assessment and Correction with Feedback}

\subsection{Long-Form Factuality Assessment}
Let $y$ be the long-form response generated by an LLM to a query $x$. We assume that $y$ can be decomposed into a set of $n$ \emph{atomic units} (or \emph{atoms}) that can be either true or false, denoted by $\cA_y = \{a_1, a_2, \ldots a_n\}$ \cite{factscore2023emnlp,song2024veriscore,wei2024longform}. An atomic unit $a_i \in \cA_y$ is defined as a short sentence conveying one piece of information (e.g., a claim or a fact). Furthermore, given an external knowledge source $\cK$\footnote{For example, $\cK$ could be Wikipedia, Google Search, or a collection of documents embedded into a vector database.}, we say that an atomic unit $a_i\in \cA_y$ is \emph{supported} by $\cK$ if there exists at least one piece of information in $\cK$ (e.g., a passage) called a \emph{context} that undebatably supports $a_i$. Otherwise, the atomic unit is \emph{not supported}.

The \emph{factual precision} denoted by $Pr(y)$ of the response $y$ with respect to a knowledge source $\cC$ is defined as: $Pr(y) = \frac{S(y)}{|\cA_y|}$, where $S(y) = \sum_{i=1}^{n} \mathbb{I}[a_i \textrm{ is supported by } \cK]$ is the number of supported atomic units. The \emph{factual recall} $R_K(y)$ up to the $K$-th supported atomic unit  is given by: $R_K(y) = \min(\frac{S(y)}{K}, 1)$. Finally, an $F_1$ measure for long-form factuality denoted by $F1@K$ can be defined as: $F_1@K(y) = \frac{2\cdot Pr(y) \cdot R_K(y)}{Pr(y) + R_K(y)}$ if $S(y) > 0$, and $0$ otherwise \cite{wei2024longform}.

\subsection{The \textsc{FactReasoner} Factuality Assessor}

\begin{figure}
    \centering
    \includegraphics[width=\linewidth]{figs/fr-pipeline.png}
    \includegraphics[width=0.8\linewidth]{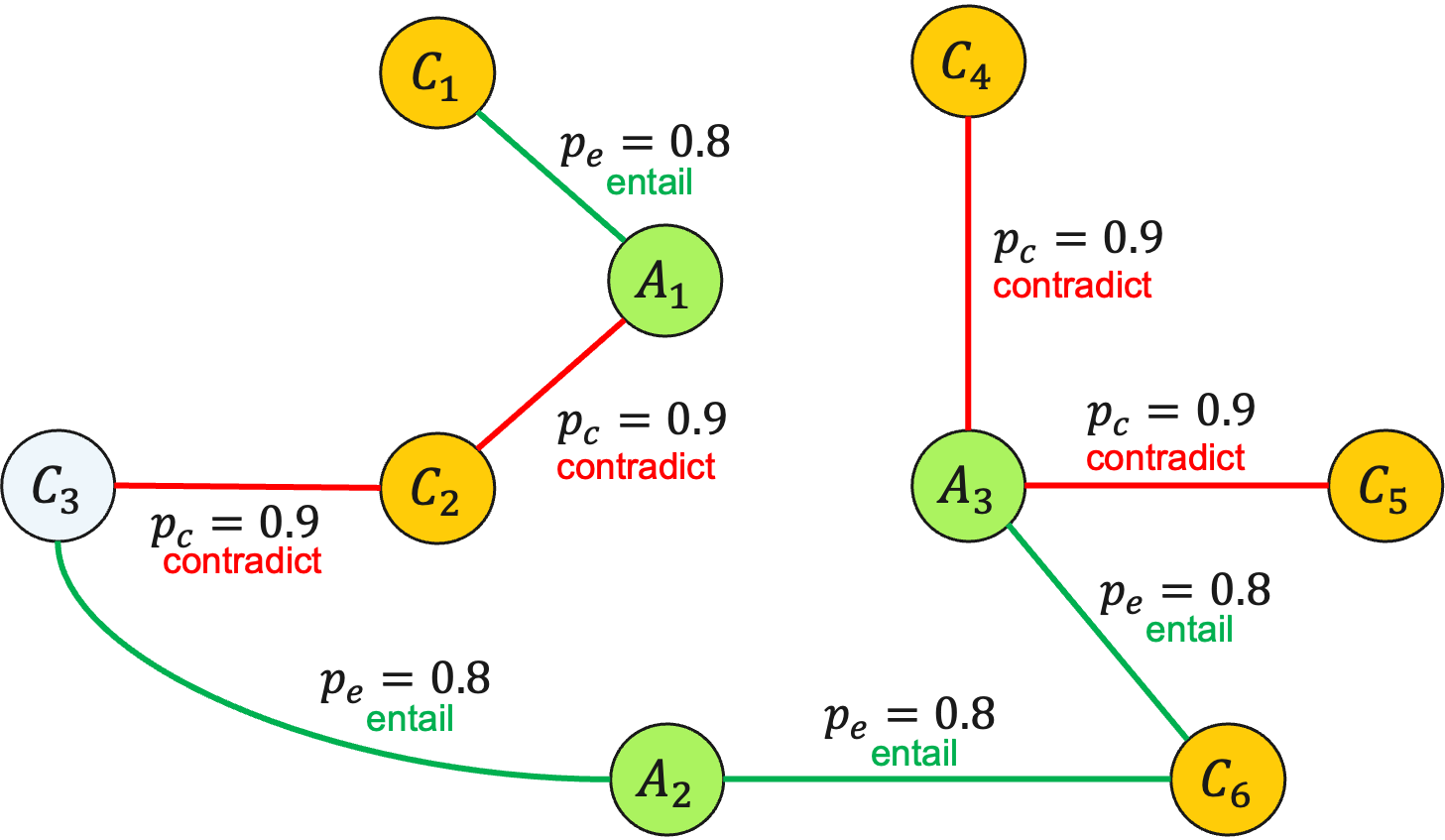}
    \caption{\textsc{FactReasoner} pipeline and an example graphical model with 3 atom and 6 context variables.}
    \label{fig:fr-pipeline}
\end{figure}

Existing long-form factuality assessors such as FactScore \cite{factscore2023emnlp}, VeriScore \cite{song2024veriscore} and others \cite{wei2024longform,bayat2025factbench} are prompt-based approaches that essentially prompt an LLM to determine if each atomic unit of the response is supported (factually correct) or not by the retrieved evidence. \textsc{FactReasoner} \cite{marinescu2025fr} is a recent factuality assessor that, unlike the prompt-based approaches, leverages probabilistic reasoning to assess the factuality of the generated response with respect to an external knowledge source $\cK$.

Specifically, the \textsc{FactReasoner} pipeline for factuality assessment shown in Figure \ref{fig:fr-pipeline} (top) consists of four main stages called \emph{Atomizer}, \emph{Reviser}, \emph{Retriever} and \emph{Evaluator}, respectively. The \emph{Atomizer} prompts an LLM to decompose the response $y$ into a set of $n$ atomic units $\cA_y$ by applying any of the decomposition strategies proposed recently \cite{factscore2023emnlp,bayat2025factbench}. Subsequently, the \emph{Reviser} also uses an LLM to revise the atoms such that the pronouns, unknown entities, or incomplete names are replaced with their corresponding named entities in the response \cite{wei2024longform}. Next, the \emph{Retriever} is responsible for querying an external knowledge source $\cK$ to retrieve the contexts relevant to the response's atoms \cite{song2024veriscore}. Finally, the \emph{Evaluator} constructs a graphical model representing a joint probability distribution over the atomic units in the response and their corresponding contexts in $\cC$. For each atom $a_i$, it then computes the posterior marginal probability $P(a_i)$ which quantifies the likelihood the $a_i$ is true (or supported) given the information available in $\cK$. Finally, the factuality measures, such as $Pr(y)$ or $F_1@K(y)$, can be readily calculated using the supported atoms \cite{marinescu2025fr}.

\paragraph{Graphical Model.} \textsc{FactReasoner}'s graphical model is defined by a tuple $\langle \bX, \bD, \bF \rangle$ where $\bX$ is a set of Boolean variables $A_i$ or $C_j$ associated with each atom $a_i \in \cA_y$ or context $c_j \in \cC_y$, respectively. The domains $\bD$ are $\{\text{true}, \text{false}\}$ for all variables (e.g., we denote $a_i$ and $\neg a_i$ for $A_i=\text{true}$ and $A_i=\text{false}$, respectively). Each variable has a unary factor $f(\cdot)$ in $\bF$ encoding prior belief. Atoms have uniform priors: $f(a_i)=f(\neg a_i)=0.5$. Contexts from reliable sources have high prior (e.g., $f(c_j)=0.99$), while less reliable sources use smaller values. Binary factors $f(A_i,C_j)$ and $f(C_j,C_k)$ capture probabilistic logical relations between utterances. A relation model $p_\theta(\cdot|t,t')$ (either a specialized BERT model or an LLM) predicts the most likely relation from \{neutral, entail, contradict\} together with its probability. Therefore, each factor $f(A_i,C_j)$ or $f(C_j,C_k)$ represents a probabilistic encoding of an entailment relationship (i.e., context $C_j$ entails or supports atom $A_i$) or a contradiction relationship (i.e., context $C_j$ contradicts atom $A_i$), respectively. Note that the neutral relationships are ignored.

\begin{example}
Figure \ref{fig:fr-pipeline} (bottom) shows an example graphical model with 3 atoms ($A_1$, $A_2$, $A_3$), 6 contexts ($C_1$, ..., $C_6$) and 8 binary relations between atoms and contexts. Note that each edge in the graph is labeled by the probability of the corresponding entailment or contradiction relationship.
\end{example}

\subsection{Post-hoc Correction with Feedback}
In general, a \emph{post-hoc correction strategy} leverages a Refine Model $\mathcal{R}$ to correct the response $\hat{y}$ generated by a Language Model $\mathcal{M}$ to a user query $x$ using the feedback provided by a Critic Model $\mathcal{C}$ regarding the quality of the response.

\begin{figure}[t]
    \centering
    \includegraphics[width=\linewidth]{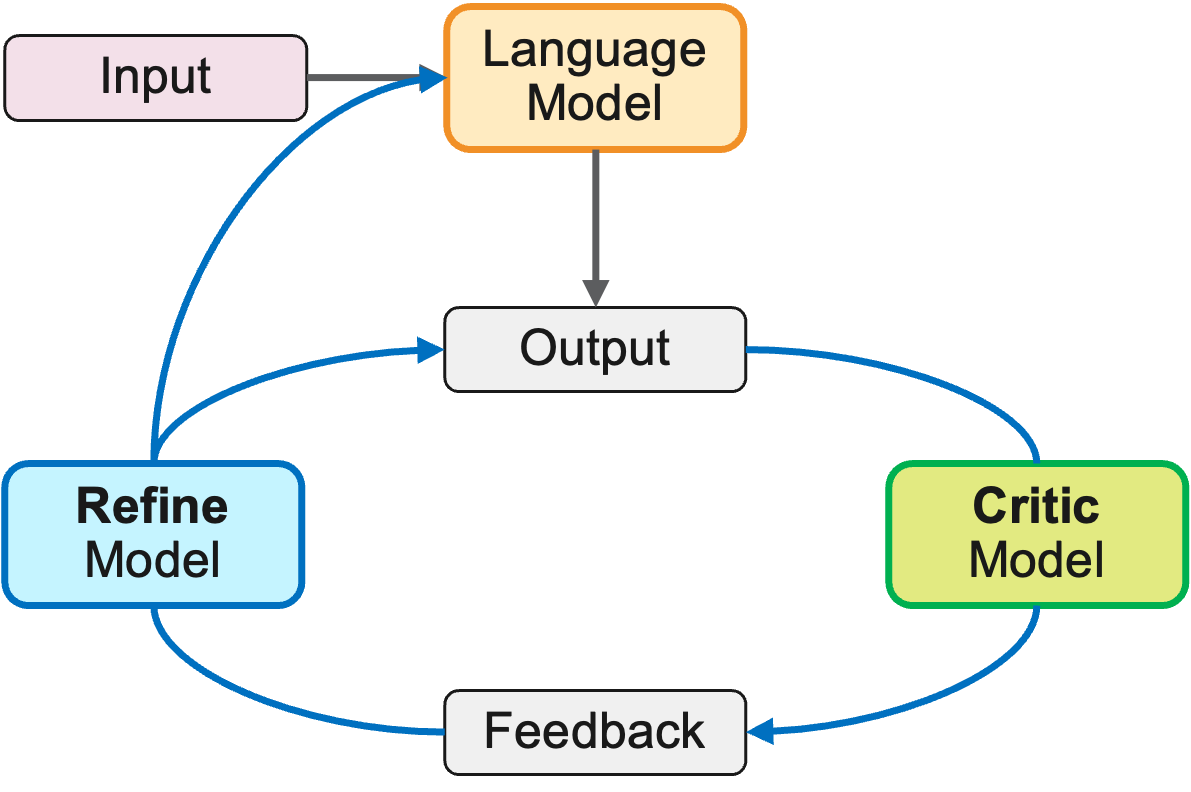}
    \caption{Post-hoc correction with feedback.}
    \label{fig:corrector}
\end{figure}

Formally, let $\mathcal{M}: \mathcal{X} \rightarrow \mathcal{Y}$ be a language model that performs a specific task by mapping the user input $x \in \mathcal{X}$ to an output text $\hat{y} \in \mathcal{Y}$. A wide range of NLP tasks adhere to this formulation, such as QA, summarization, translation, and more. The initial generation $\hat{y}$ may have problems such as hallucinations, factual errors, or incorrect reasoning. The critic model $\mathcal{C}: \mathcal{X} \times \mathcal{Y} \rightarrow \mathcal{F}$ learns to generate feedback $x,\hat{y} \rightarrow c$ where $\hat{y} \sim \mathcal{M}(x)$ is the output of the language model, and $c$ is the feedback of some format such as a scalar value or natural language. A simple example is binary feedback of whether
the output is good or bad given the input $\mathcal{C}: \mathcal{X} \times \mathcal{Y} \rightarrow \{0,1\}$. The refine model $\cR: \cX \times \cY \times \cF \rightarrow \cY$ learns to repair/correct an output $x,\hat{y},c \rightarrow y_{new}$ based on the feedback $c$, where $y_{new}$ is the revised output \cite{pan2024survey}.

Figure \ref{fig:corrector} shows how the components of a post-hoc correction pipeline interact to repair the output of a language model.

\section{Additional Details on Experiments}
\label{sec-experiments-all}
In this section, we report additional details on our experiments.

\subsection{Baseline Correctors}
For our purpose, we consider two recent state-of-the-art post-hoc correction with feedback approaches, namely CRITIC \cite{gou2024critic} and RAC \cite{li2024rac}. Specifically, CRITC is a prompt-based iterative correction method that uses LLMs with retrieved knowledge. At each iteration, CRITIC prompts an LLM to generate a query to retrieve relevant information from a knowledge base and subsequently revise the output based on the history of revising and decide the most possible answer. RAC is a post-hoc factuality correction method designed to improve the accuracy of LLMs responses without additional fine-tuning. Its key steps are: the model’s generated text is broken down into individual factual statements, relevant information is fetched from trusted sources (e.g., Wikipedia or document collections), and each atomic fact is checked against retrieved evidence using a verification model. Incorrect statements are then corrected based on reliable content.

In addition, we consider two prompting strategies, denoted LLM1 and LLM2. The first, LLM1, performs response correction using only the model’s internal knowledge, without incorporating any retrieved context. In contrast, LLM2 uses the contexts retrieved for the question only and instructs the model to ignore its internal knowledge to refine the response (see Appendix \ref{sec-prompts} for the actual prompts).

In our experiments, we instantiated the competing correctors with open-source LLMs belonging to the IBM Granite\footnote{\texttt{https://huggingface.co/ibm-granite}}, Meta LLaMA\footnote{\texttt{https://huggingface.co/meta-llama}},  MistralAI Mixtral\footnote{\texttt{https://huggingface.co/mistralai}} and OpenAI\footnote{\texttt{https://openai.com}} families, namely: granite-4.0-h-small, llama-3.3-70b-instruct, mixtral-8x22b-instruct, and gpt-oss-120b, respectively. All our LLMs are hosted remotely on compute nodes with A100 80GB GPUs and accessed via \texttt{litellm} APIs capable of serving 1500 prompts per second.

\paragraph{External Knowledge Source.} We consider Google Search results as our external knowledge source $\cK$. For a given atom, the top $k$ results are retrieved as contexts from google.com using the Serper API\footnote{\texttt{https://serper.dev}}. In this case, a context is a tuple $(t,l,s,d)$, where $t$ is the title of the web-page, $l$ is the link, $s$ is a short text snippet or summary and $d$ is the content retrieved from $l$ (but capped at max 4000 characters). We used $k=3$ for the Google Search results \cite{factscore2023emnlp,wei2024longform}.

\subsection{Datasets}
We experimented with the following datasets: a reduced version of the \textsc{Veli5} dataset presented in the previous section and consisting of 200 instances sampled uniformly at random across all categories,  Biographies (\textsc{Bio}) \cite{factscore2023emnlp}, and AskHistorians (\textsc{AskHist}) \cite{xu2023}. These datasets have been widely adopted in prior work and are considered representative benchmarks for long-form factuality assessment, as they encompass a diverse range of topics and tasks, including creative writing, history, astronomy, chemistry, and more.

The \textsc{Bio} dataset contains 183 biographic passages spanning up to two paragraphs that were generated by ChatGPT for various person entities that have a Wikipedia page \cite{factscore2023emnlp}. The \textsc{AskHist} datasets contains 200 questions sourced from the r/AskHistorians Reddit forum together with a long-form response of up to two paragraphs in length generated using the llama-3.3-70b-instruct model \cite{marinescu2025fr}.

Additionally, we include results on the \textsc{Conflicts} dataset introduced in \cite{marinescu2025fr}. This dataset comprises 100 claims (atomic units) randomly drawn from the ConflictBank benchmark \cite{su2024conflictbank}. Each claim, originally sourced from Wikidata, is assumed to be true (i.e., supported). For our purpose, we treat each claim as a \emph{response} and provide both a supporting context (as in ConflictBank) and a conflicting context (representing misinformation). Importantly, since all responses are true, they require no corrections, offering a controlled setting for our evaluation.

\subsection{Measures of Performance}
For each dataset $\mathcal{D}$ and each competing corrector, we compute three factuality metrics: precision ($Pr$), recall at $K$ ($R@K$), and $F_1@K$, averaged over all prompts in $\mathcal{D}$, where $K$ is set to the median number of atoms. These metrics are assessed using the \textsc{FactReasoner} framework \cite{marinescu2025fr}. In addition, we report two complementary measures: verifiability ($V$) and comprehensiveness ($C$). The verifiability score $V$ is defined as the number of atoms in the response (or correction) that can be verified—i.e., atoms connected to either supporting or contradicting evidence in FR’s graph. Comprehensiveness $C$ quantifies coverage and is given by $C = \frac{|\cA_{in}|}{|\cA_{in}| + |\cA_{out}|}$, where $\cA_{in}$ denotes the atoms covered by the response (or correction), and $\cA_{out}$ denotes atoms that are \emph{uncovered} or missing \cite{dejl2025compr}.

For each score $S$ (e.g., precision), we also report its \emph{relative gain}, denoted as $G(S)$ and defined by:
\[
G(S) = \frac{2 \cdot (S_c - S_r)}{S_c + S_r}
\]
\noindent where $S_r$ and $S_c$ are the scores of the original response and the correction, respectively. A positive $G(S)$ indicates that the correction outperforms the response, while a negative value means that the correction performs worse. By construction, $G(S)$ ranges from $-2$ to $2$ and remains well defined even when either $S_r$ or $S_c$ equals zero.

\subsection{Detailed Results}

\paragraph{Number of Atoms.} Table \ref{tab:atoms-all} presents the mean number of atoms and their standard deviations for both initial responses and subsequent corrections generated by the different correctors across datasets and base models. Notably, in some cases, the corrections contain substantially more atoms than the original responses. This increase may contribute to the observed decline in relative gains for factuality metrics, including the negative gains reported in Tables \ref{tab:veli5-gains-all}, \ref{tab:bio-gains-all}, and \ref{tab:askhist-gains-all}.

\paragraph{\textsc{Conflicts}.} Table \ref{tab:conflicts-all} summarizes the results obtained on our controlled experiment with the \textsc{Conflicts} dataset. In this setting, since the original is known to be true and doesn't require correction, we assess the similarity between correction and response using the ROUGE, BLEU, and BLEURT metrics \cite{lin2004rouge,papineni2002bleu,sellam2020bleurt}. Additionally, we report JUDGE, which counts the number of instances where an LLM-as-a-Judge model (in our case DeepSeek-3.2), prompted appropriately, infers that the correction is equivalent to the original response. The table shows the mean and standard deviation for all metrics. The results indicate that \textsc{FactCorrector} achieves superior performance with the Mixtral and Granite models, while ranking second when using LLaMA and GPT-OSS models, where RAC attains the best scores. Therefore, both \textsc{FactCorrector} and RAC correctly identify that no correction is needed. In contrast, the LLM1 and LLM2 baselines perform rather poorly almost always producing corrections significantly different from the response, while CRITIC often struggles to follow its instructions also leading to poor results.

\paragraph{\textsc{Veli5}.} In Table \ref{tab:veli5-gains-all} we report the mean relative gains and standard deviations obtained for the factuality metrics on the \textsc{Veli5} dataset using the mixtral-8x22b-instruct, llama-3.3-70b-instruct, granite-4.0-h-small and gpt-oss-120b models. The results demonstrate that \textsc{FactCorrector} consistently delivers strong and reliable improvements in factuality across all evaluated metrics and models. For example, when considering relative gains on precision and F1@K, FC outperforms all baselines, achieving macro-averaged gains of 0.315 for precision, and 0.2325 for F1@K — substantially higher than the next-best corrector (see also Figure \ref{fig:veli5-gains-macro}).  Importantly, FC is often best or tied-best in many metric–model combinations, showing strong generalization across diverse LLM architectures. LLM1 and LLM2 show strong performance in this case, likely due to the provenance of the VELI5 questions which were originally derived from public Reddit forums. Therefore, it is plausible that similar content was included in these models’ pre-training corpora which may confer a significant advantage on this dataset.

\begin{figure}[t!]
    \centering
    \includegraphics[width=\linewidth]{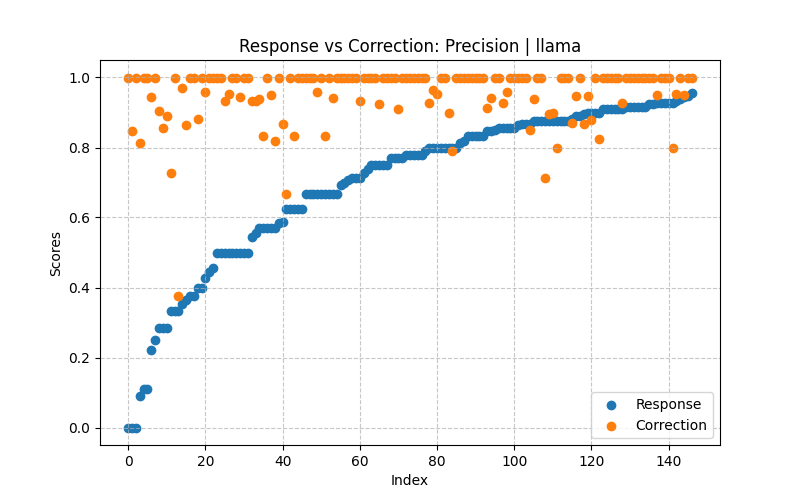}
    \includegraphics[width=\linewidth]{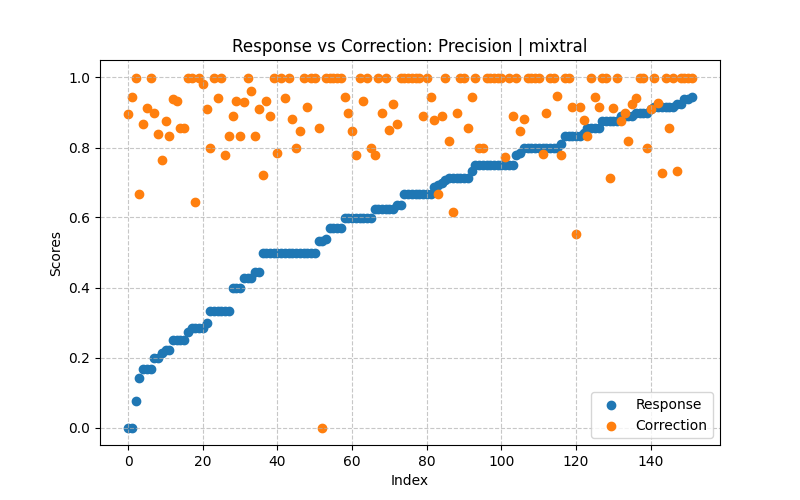}
    \includegraphics[width=\linewidth]{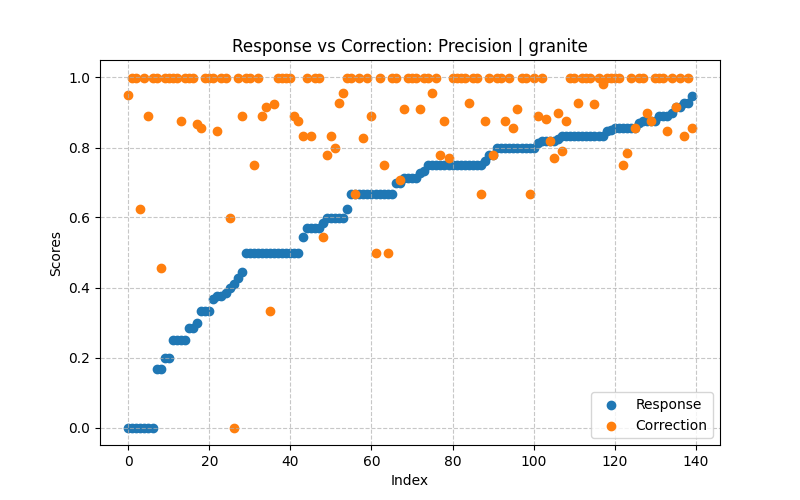}
    \includegraphics[width=\linewidth]{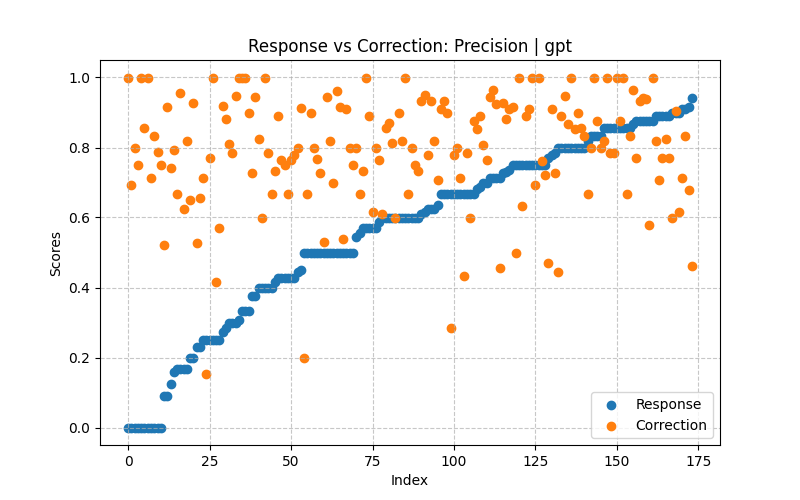}
    \caption{Factual precision: response vs correction by \textsc{FactCorrector} on the \textsc{Veli5} dataset.}
    \label{fig:fs-veli5}
\end{figure}

Table \ref{tab:veli5-precision-all} reinforces these findings by examining factual precision before and after correction. FC achieves the highest post-correction precision on two out of four models (Llama, Mixtral) and ranks second on Granite and GPT-OSS, where it still delivers large relative gains. On average, FC improves precision by +0.21, outperforming all baselines, including LLM1 (+0.17) and RAC (+0.16). While RAC slightly surpasses FC on GPT-OSS in final precision, FC dominates in overall gains, suggesting a more robust correction strategy. In contrast, CRITIC exhibits inconsistent behavior, even reducing precision for GPT-OSS. Finally, Figure \ref{fig:fs-veli5} plots the factual precision of both the original response and its correction generated by \textsc{FactCorrector} highlighting the substantial positive gains achieved by the corrector.

\begin{figure}[t]
    \centering
    \includegraphics[width=0.9\linewidth]{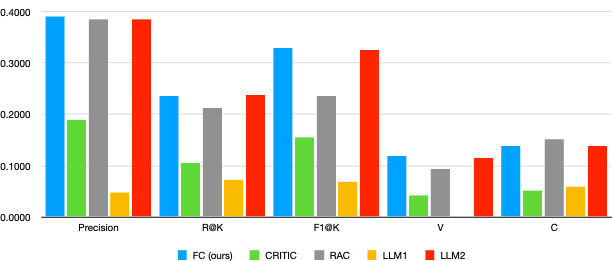}
    \caption{Mean relative gains (macro) for factuality metrics across models on the \textsc{Bio} dataset.}
    \label{fig:bio-gains-macro}
\end{figure}

\paragraph{\textsc{Bio}.} Table \ref{tab:bio-gains-all} presents the mean relative gains and standard deviations obtained on the \textsc{Bio} dataset. FC exhibits robust performance across models, achieving its highest gains with Mixtral and Granite, and ranking second for Llama. In contrast, LLM1 performs poorly, particularly on Granite and GPT-OSS. Both RAC and LLM2 remain highly competitive on this dataset, while CRITIC, as observed previously, struggles to follow instructions, leading to inconsistent results. Figure \ref{fig:bio-gains-macro} further illustrates the macro-gains across models, showing that FC slightly outperforms RAC and LLM2. Notably, the \textsc{Bio} dataset is sourced from Wikipedia \cite{factscore2023emnlp} — a common component of the pre-training corpora for most modern LLMs, including those evaluated in our experiments. This explains the strong performance of RAC and LLM2 which rely heavily on their internal knowledge. Looking at Table \ref{tab:bio-precision-all} which reports the mean factual precision before and after correction we can see that FC is the best performing corrector on two models (Mixtral, Granite) and second best on Llama. In contrast, GPT-OSS seems to be better suited for RAC and LLM2.

\paragraph{\textsc{AskHist}.} Table \ref{tab:askhist-gains-all} summarizes the mean relative gains and standard deviations obtained on the \textsc{AskHist} dataset. Unlike previous experiments, the gains here are considerably smaller. This is primarily because the factual precision of the original responses is already high—typically exceeding 89\% on average—leaving limited room for improvement through correction. Interestingly, all other baselines exhibit negative gains, indicating that their corrections are less factual than the original responses. In contrast, FC maintains robust performance across all models, almost always producing corrections that are at least as factual as the original responses and, in many cases, substantially better. The results in Table \ref{tab:askhist-precision-all} corroborate this observation and further validate the effectiveness of our proposed FC corrector.

In summary, \textsc{FactCorrector} (FC) consistently delivers the most reliable improvements in factuality, across all datasets. Specifically, on \textsc{Veli5}, FC achieves substantial gains across precision, recall@K, and F1@K, outperforming all baselines and demonstrating strong generalization across models. For \textsc{Bio}, FC remains robust, slightly surpassing RAC and LLM2 despite their advantage from Wikipedia-based pre-training, and achieves top or near-top factual precision on most models. On \textsc{AskHist}, where original responses are already highly factual, FC still maintains positive gains, unlike other baselines that often degrade performance. These results highlight FC’s effectiveness and adaptability across diverse datasets and model architectures.

\paragraph{Statistical Significance Tests.}
We performed one-sided t-tests on the mean relative gain in precision achieved by FC compared to competing methods. The corresponding ppp-values are reported in Table \ref{tab:p-values}. The near-zero ppp-values for the precision metric provide strong statistical evidence that FC consistently and significantly outperforms all baselines across datasets and underlying model configurations. These results underscore the robustness of FC and its ability to deliver substantially more accurate factual corrections than existing approaches.

\subsection{Results for SFT Correction}
We used the larger \textsc{Veli5} dataset to train a LoRA adapter on the Granite-Guardian-5B model for generating corrections. The hyperparameters of the LoRA adapter were the following: $r$ = 64, $\alpha$ = 128, bias = none, lr = 1e-5, schedule = cosine, batch size = 16 and we used all linear layers.

Table \ref{tab:sft-veli5} reports the mean relative gains on the \textsc{Veli5} dataset when using the \textsc{FactCorrector} pipeline versus the LoRA-based correction approach (denoted as SFT). Overall, the SFT method produces substantially better corrections than the original responses, which is expected given that the evaluation instances originate from the same distribution as the training data.

Tables \ref{tab:sft-bio} and \ref{tab:sft-askhist} present results on the \textsc{Bio} and \textsc{AskHist} datasets, which are out-of-distribution relative to the SFT training data. As expected, the gains achieved by SFT are smaller than before; however, most gains remain non-negative, indicating that the SFT approach generalizes reasonably well. This is noteworthy because SFT is significantly more cost-efficient, requiring only a single LLM call per correction, compared to the more complex \textsc{FactCorrector} pipeline, which involves multiple LLM calls for each instance.

\subsection{Human Evaluation of \textsc{Veli5}}
\label{sec-human-eval}
We conducted a human evaluation to assess the quality and correctness of VELI5 corrections. Annotators were presented with the original question, the LLM response, a highlighted atom text extracted from the response, relevant contextual information, and the correction. For each instance, annotators made three judgments: (1) whether the incorrect atom had been appropriately corrected given the provided context (labeled as corrected, undercorrected (which also includes not corrected and wrongly corrected), rightly overcorrected, or invalid); (2) whether the factual content of a correct atom was preserved in the correction (labeled as preserved, partially preserved, not preserved, or invalid), and separately rate each atom’s relevance to the question as Relevant, Partially relevant, Irrelevant, Invalid (hallucinated atom), or Invalid (other reasons). Clear annotation guidelines and examples were provided to ensure consistency. This evaluation setup allowed us to measure both factual alignment with external context and faithfulness to the original atomic information, providing a fine-grained assessment of correction quality beyond overall response accuracy.

We design a Label Studio interface to evaluate revisions at multiple granularities. For incorrect or unverified response atoms, annotators judge whether each atom is Corrected, Undercorrected, Overcorrected, Invalid (hallucinated atom), or Invalid (other reasons) (with a free-text rationale when “Invalid (other reasons)” is selected), and separately rate each atom’s relevance to the question as Relevant, Partially relevant, Irrelevant, Invalid (hallucinated atom), or Invalid (other reasons) (again with an optional rationale for the latter). For correct response atoms, annotators assess information preservation in the correction as Preserved, Partially preserved, Not preserved, Invalid (hallucinated atom), or Invalid (other reasons) (with rationale text for “Invalid (other reasons)”), and also label each correct atom’s relevance to the question using the same five-way relevance scale. Finally, at the response level, annotators label model-response relevance and correction relevance to the question as Relevant, Partially Relevant, or Irrelevant, providing an end-to-end check that the revision remains on-topic while atom-level labels capture whether factual errors were fixed and correct content was retained.

\begin{table}[t]
\centering
\small

\begin{tabular}{l c}
\toprule
\textbf{Metric} & \textbf{\%} \\
\midrule
\multicolumn{2}{l}{\textbf{Dataset Summary}} \\
Tasks (filtered) & 30 \\
Incorrect atoms / task & 2.30 $\pm$ 1.39 \\
Correct atoms / task & 4.40 $\pm$ 2.66 \\
\midrule
\multicolumn{2}{l}{\textbf{Task-level relevance}} \\
Response relevant & 90.0\% \\
Response partially relevant & 10.0\% \\
Response irrelevant & 0.0\% \\
Correction relevant & 96.7\% \\
Correction partially relevant & 3.3\% \\
Correction irrelevant & 0.0\% \\
\midrule
\multicolumn{2}{l}{\textbf{Incorrect atoms --- correction}} \\
Corrected & 67.4 $\pm$ 38.5 \\
Undercorrected & 26.4 $\pm$ 36.8 \\
Overcorrected & 3.3 $\pm$ 18.3 \\
Invalid & 2.9 $\pm$ 12.6 \\
\midrule
\multicolumn{2}{l}{\textbf{Incorrect atoms --- relevance}} \\
Relevant & 75.0 $\pm$ 43.1 \\
Partially relevant & 3.3 $\pm$ 18.3 \\
Irrelevant & 5.0 $\pm$ 20.1 \\
\midrule
\multicolumn{2}{l}{\textbf{Correct atoms --- preservation}} \\
Preserved & 60.9 $\pm$ 34.9 \\
Partially preserved & 9.2 $\pm$ 14.8 \\
Not preserved & 27.7 $\pm$ 29.5 \\
Invalid & 2.3 $\pm$ 5.3 \\
\midrule
\multicolumn{2}{l}{\textbf{Correct atoms --- relevance}} \\
Relevant & 72.4 $\pm$ 40.7 \\
Partially relevant & 7.9 $\pm$ 17.1 \\
Irrelevant & 3.0 $\pm$ 13.2 \\
\bottomrule
\end{tabular}

\caption{Atom-level evaluation and task-level relevance (percent mean $\pm$ std).}
\label{tab:atom-eval-all}
\end{table}

Table \ref{tab:atom-eval-all} summarizes the results of our human evaluation conducted on 30 filtered tasks (or instances) from our \textsc{Veli5} dataset. Each task comprises on average 2.30 incorrect atoms and 4.40 correct atoms. At the task level, the original responses demonstrated high relevance with respect to the user question, with 90\% classified as fully relevant and 10\% as partially relevant, while corrections achieved even greater relevance to the question, with 96.7\% fully relevant. For incorrect atoms, 67.4\% were successfully corrected by \textsc{FactCorrector}, though 26.4\% remained undercorrected, and only minor proportions were overcorrected (3.3\%) or invalid (2.9\%). Correct atoms in the original response were preserved by the correction in 60.9\% of cases, with 27.7\% not preserved and small fractions partially preserved or invalid. Relevance analysis indicated that both incorrect and correct atoms were predominantly relevant to the user question (75.0\% and 72.4\%, respectively), underscoring the overall effectiveness of the correction process.

\begin{table}[t]
    \centering
    \resizebox{\linewidth}{!}{
    \begin{tabular}{l|c|c|c|c}
    corrector & ROUGE $\uparrow$ & BLEU $\uparrow$ & BLEURT $\uparrow$ & JUDGE $\uparrow$ \\
    \toprule
    \multicolumn{5}{c}{\texttt{mixtral-8x22b-instruct}}\\
    \midrule
    CRITIC &  0.14$\pm$0.07   & 0.02$\pm$0.03   & -0.69$\pm$0.25   & 0.41  \\
    RAC    &  \underline{0.54}$\pm$0.23   & 0.23$\pm$0.27   &  0.29$\pm$0.43   & \underline{{\bf 0.87}}  \\
    LLM1   &  0.15$\pm$0.05    & 0.02$\pm$0.02   & -0.63$\pm$0.21   & 0.30  \\
    LLM2   &  0.20$\pm$0.10    & 0.05$\pm$0.05   & -0.48$\pm$0.22   & 0.10  \\
    \midrule
    FC (ours) & {\bf 0.89$\pm$0.26} & {\bf 0.87$\pm$0.32} & {\bf 0.73$\pm$0.47} & {\bf 0.87} \\
    \midrule
    \multicolumn{5}{c}{\texttt{llama-3.3-70b-instruct}}\\
    \midrule
    CRITIC &  0.32$\pm$0.27   &  0.15$\pm$0.28  &  -0.36$\pm$0.56   & 0.31  \\
    RAC    &  {\bf 0.87$\pm$0.20}   &  {\bf 0.77$\pm$0.32}  &  {\bf 0.65$\pm$0.43}   & {\bf 0.77}  \\
    LLM1   &  0.19$\pm$0.11   &  0.03$\pm$0.05  & -0.62$\pm$0.33    & 0.17  \\
    LLM2   &  0.21$\pm$0.10   &  0.04$\pm$0.04  & -0.46$\pm$0.22    & 0.06  \\
    \midrule
    FC (ours) & \underline{0.73}$\pm$0.33 & 0.62$\pm$0.45   & 0.46$\pm$0.56   & \underline{0.60} \\
    \midrule
    \multicolumn{5}{c}{\texttt{granite-4.0-h-small}}\\
    \midrule
    CRITIC &  0.15$\pm$0.11   & 0.03$\pm$0.10   & -0.69$\pm$0.26   & 0.20  \\
    RAC    &  \underline{0.74}$\pm$0.23   & 0.48$\pm$0.37   & 0.49$\pm$0.46   &  \underline{0.73} \\
    LLM1   &  0.15$\pm$0.07   & 0.02$\pm$0.02   & -0.68$\pm$0.23   & 0.17  \\
    LLM2   &  0.42$\pm$0.24   & 0.15$\pm$0.17   & -0.10$\pm$0.44   & 0.13  \\
    \midrule
    FC (ours) & {\bf 0.83$\pm$0.30} & {\bf 0.76$\pm$0.39}   & {\bf 0.58$\pm$0.57}   & {\bf 0.78} \\
    \midrule
    \multicolumn{5}{c}{\texttt{gpt-oss-120b}}\\
    \midrule
    CRITIC &  0.12$\pm$0.07   & 0.01$\pm$0.01   & -0.89$\pm$0.22   & 0.09  \\
    RAC    &  {\bf 0.85$\pm$0.17}   & {\bf 0.60$\pm$0.33}   & {\bf 0.74$\pm$0.29}   & {\bf 0.89}  \\
    LLM1   &  0.08$\pm$0.03  & 0.00$\pm$0.01   & -0.83$\pm$0.15   & 0.37  \\
    LLM2   &  0.16$\pm$0.06  & 0.02$\pm$0.02   & -0.54$\pm$0.18   & 0.19  \\
    \midrule
    FC (ours) & \underline{0.72}$\pm$0.37 & 0.66$\pm$0.46   & 0.44$\pm$0.67   & \underline{0.66} \\
    \midrule
    \multicolumn{5}{c}{\texttt{granite-4.0-h-tiny}}\\
    \midrule
    CRITIC & 0.55$\pm$0.43  & 0.49$\pm$0.47 & 0.08$\pm$0.81  & 0.77 \\
    RAC    & \underline{0.83}$\pm$0.18  & \underline{0.59}$\pm$0.33 & \underline{0.67}$\pm$0.36   & \underline{0.84} \\
    LLM1   & 0.22$\pm$0.19  & 0.08$\pm$0.19 & -0.44$\pm$0.40  & 0.40 \\
    LLM2   & 0.38$\pm$0.29  & 0.15$\pm$0.21 & -0.23$\pm$0.47  & 0.29 \\
    \midrule
    FC (ours) & {\bf 0.97$\pm$0.15} & {\bf 0.95$\pm$0.19} & {\bf 0.86$\pm$0.28} & {\bf 0.97} \\
    \midrule
    \end{tabular}
    }
    \caption{Results obtained on the \textsc{Conflicts} dataset.}
    \label{tab:conflicts-all}
\end{table}

\begin{table*}[t!]
    \centering
    \small
    \begin{tabular}{l|c|c|c|c|c}
    corrector &  Pr $\uparrow$   & R@K $\uparrow$ & F1@K $\uparrow$ & V $\uparrow$ & C $\uparrow$ \\
\toprule
\multicolumn{6}{c}{\texttt{mixtral-8x22b-instruct}} \\
\midrule
CRITIC  & 0.26$\pm$0.49 & 0.08$\pm$0.33  & 0.19$\pm$0.41  & 0.09$\pm$0.32   &  0.06$\pm$0.18 \\
RAC     & 0.24$\pm$0.49 & 0.08$\pm$0.33  & 0.18$\pm$0.41  & 0.08$\pm$0.29   &  0.06$\pm$0.20 \\
LLM1    & 0.32$\pm$0.45 & 0.10$\pm$0.30  & 0.23$\pm$0.37  & 0.12$\pm$0.30   &  0.07$\pm$0.17 \\
LLM2    & 0.32$\pm$0.45 & 0.10$\pm$0.30  & 0.23$\pm$0.38  & 0.13$\pm$0.31   &  0.07$\pm$0.17 \\
\midrule
FC (ours)  & {\bf 0.34$\pm$0.46} & 0.09$\pm$0.34  & {\bf 0.24$\pm$0.40}  & {\bf 0.14$\pm$0.29}   &  {\bf 0.07$\pm$0.16} \\
SFT (ours) & {\bf 0.19$\pm$0.51} & 0.10$\pm$0.41  & {\bf 0.16$\pm$0.45}  & {\bf 0.05$\pm$0.33}   &  {\bf 0.08$\pm$0.23} \\

\midrule
\multicolumn{6}{c}{\texttt{llama-3.3-70b-instruct}} \\
\midrule
CRITIC  & 0.19$\pm$0.41 & 0.06$\pm$0.29  & 0.13$\pm$0.34  & 0.03$\pm$0.17   &  0.05$\pm$0.25 \\
RAC     & 0.13$\pm$0.37 & 0.02$\pm$0.22  & 0.09$\pm$0.29  & 0.02$\pm$0.08   &  0.05$\pm$0.20 \\
LLM1    & 0.25$\pm$0.40 & 0.06$\pm$0.29  & 0.18$\pm$0.34  & 0.05$\pm$0.16   &  0.08$\pm$0.24 \\
LLM2    & 0.26$\pm$0.40 & 0.07$\pm$0.29  & 0.18$\pm$0.34  & 0.05$\pm$0.16   &  0.08$\pm$0.24 \\
\midrule
FC (ours)  & {\bf 0.27$\pm$0.39} & {\bf 0.07$\pm$0.29}  & {\bf 0.18$\pm$0.34}  & {\bf 0.05$\pm$0.16}   &  {\bf 0.08$\pm$0.23} \\
SFT (ours) & {\bf 0.25$\pm$0.49} & {\bf 0.11$\pm$0.41}  & {\bf 0.19$\pm$0.44}  & {\bf 0.10$\pm$0.32}   &  {\bf 0.08$\pm$0.23} \\

\midrule
\multicolumn{6}{c}{ \texttt{granite-4.0-h-small}} \\
\midrule
CRITIC  & 0.17$\pm$0.55 & 0.04$\pm$0.43  & 0.12$\pm$0.49  & 0.05$\pm$0.38   &  0.06$\pm$0.23 \\
RAC     & 0.20$\pm$0.49 & 0.09$\pm$0.35  & 0.16$\pm$0.41  & 0.06$\pm$0.26   &  0.07$\pm$0.23 \\
LLM1    & {\bf 0.31$\pm$0.48} & {\bf 0.11$\pm$0.40}  & {\bf 0.23$\pm$0.44}  &  {\bf 0.11$\pm$0.31}   &  {\bf 0.08$\pm$0.22} \\
LLM2    & 0.20$\pm$0.53 & 0.10$\pm$0.42  & 0.16$\pm$0.47  & 0.09$\pm$0.32   &  0.05$\pm$0.25 \\
\midrule
FC (ours)  & 0.29$\pm$0.49 & 0.10$\pm$0.43  & 0.21$\pm$0.45  & 0.11$\pm$0.32   &  0.07$\pm$0.23 \\
SFT (ours) & 0.22$\pm$0.50 & 0.11$\pm$0.40  & 0.18$\pm$0.45  & 0.09$\pm$0.32   &  0.07$\pm$0.24 \\
\midrule
\multicolumn{6}{c}{\texttt{gpt-oss-120b}} \\
\midrule
CRITIC  & -0.04$\pm$0.62 & 0.01$\pm$0.54  & -0.02$\pm$0.57  & -0.01$\pm$0.30   &  -0.03$\pm$0.33 \\
RAC     & 0.34$\pm$0.60 & 0.15$\pm$0.48  & 0.27$\pm$0.53  & 0.06$\pm$0.31   &  0.13$\pm$0.34 \\
LLM1    & 0.21$\pm$0.64 & 0.16$\pm$0.51  & 0.19$\pm$0.57  & -0.01$\pm$0.30   &  0.13$\pm$0.34 \\
LLM2    & 0.33$\pm$0.64 & 0.20$\pm$0.52  & 0.28$\pm$0.57  & 0.05$\pm$0.29   &  0.13$\pm$0.34 \\
\midrule
FC (ours)  & {\bf 0.36$\pm$0.58} & {\bf 0.19$\pm$0.51}  & {\bf 0.30$\pm$0.54}  & {\bf 0.07$\pm$0.26}   &  0.14$\pm$0.33 \\
SFT (ours) & {\bf 0.08$\pm$0.57} & {\bf 0.08$\pm$0.45}  & {\bf 0.08$\pm$0.51}  & {\bf 0.07$\pm$0.34}   &  0.02$\pm$0.26 \\
\midrule

    \end{tabular}
    \caption{Relative Gains for the factuality metrics obtained on the \texttt{VELI5} dataset. We show the mean and standard deviation as well as highlight the best performance.}
    \label{tab:veli5-gains-all}
\end{table*}

\begin{table*}
    \centering
    \small

    \begin{tabular}{l|cc|cc|cc|cc}
    corrector &  \multicolumn{2}{c|}{llama-3.3-70b-instruct} & \multicolumn{2}{c|}{mixtral-8x22b-instruct} & \multicolumn{2}{c|}{granite-4.0-h-small} & \multicolumn{2}{c}{gpt-oss-120b} \\
    & before & after & before & after & before & after & before & after \\
\toprule
CRITIC  & 0.78$\pm$0.24 & 0.90$\pm$0.15 & 0.70$\pm$0.25  & 0.88$\pm$0.18   &  0.74$\pm$0.26 & 0.86$\pm$0.24 & 0.62$\pm$0.28 & 0.58$\pm$0.26\\
RAC  & 0.78$\pm$0.24 & 0.91$\pm$0.14  & 0.70$\pm$0.25  & 0.86$\pm$0.18   &  0.74$\pm$0.26 & 0.88$\pm$0.17 & 0.62$\pm$0.28 & {\bf 0.85$\pm$0.20} \\
LLM1   & 0.78$\pm$0.24 & 0.95$\pm$0.07  & 0.70$\pm$0.25  & 0.91$\pm$0.13   &  0.74$\pm$0.26 & {\bf 0.94$\pm$0.10} & 0.62$\pm$0.28 & 0.72$\pm$0.18 \\
LLM2  & 0.78$\pm$0.24 & 0.96$\pm$0.07  & 0.70$\pm$0.25  & 0.91$\pm$0.11   &  0.74$\pm$0.26 & 0.86$\pm$0.17 & 0.62$\pm$0.28 & 0.79$\pm$0.18 \\
\midrule
FC (ours)  & 0.78$\pm$0.24 & {\bf 0.97$\pm$0.07}  & 0.70$\pm$0.25  & {\bf 0.93$\pm$0.11}   &  0.74$\pm$0.26 & 0.93$\pm$0.13 & 0.62$\pm$0.28 & 0.82$\pm$0.16 \\
\midrule
\end{tabular}

\caption{Mean factual precision and standard deviation before and after correction on the \textsc{VELI5} dataset.}
\label{tab:veli5-precision-all}
\end{table*}

\begin{table*}[t!]
    \centering
    \small
    \begin{tabular}{l|c|c|c|c|c}
    corrector &  Pr $\uparrow$   & R@K $\uparrow$ & F1@K $\uparrow$ & V $\uparrow$ & C $\uparrow$ \\
\toprule
\multicolumn{6}{c}{\texttt{mixtral-8x22b-instruct}} \\
\midrule
CRITIC  & 0.27$\pm$0.83 & 0.22$\pm$0.80 & 0.25$\pm$0.82  & 0.05$\pm$0.32  & 0.05$\pm$0.41  \\
RAC     & 0.39$\pm$0.68 & 0.29$\pm$0.68 & 0.34$\pm$0.68  & 0.05$\pm$0.38  & 0.13$\pm$0.35  \\
LLM1    & 0.27$\pm$0.66 & 0.25$\pm$0.64 & 0.27$\pm$0.65  & 0.02$\pm$0.33  & 0.07$\pm$0.33  \\
LLM2    & 0.42$\pm$0.72 & 0.30$\pm$0.72 & 0.37$\pm$0.72  & 0.11$\pm$0.29  & 0.07$\pm$0.41  \\
\midrule
FC (ours)  & {\bf 0.46$\pm$0.70} & {\bf 0.33$\pm$0.72}  & {\bf 0.41$\pm$0.70}  & {\bf 0.12$\pm$0.27}   &  {\bf 0.11$\pm$0.36} \\
SFT (ours) & {\bf 0.05$\pm$0.42} & {\bf 0.00$\pm$0.34}  & {\bf 0.02$\pm$0.38}  & {\bf 0.05$\pm$0.27}   &  {\bf 0.02$\pm$0.11} \\
\midrule
\multicolumn{6}{c}{\texttt{llama-3.3-70b-instruct}} \\
\midrule
CRITIC  & 0.37$\pm$0.61 & 0.18$\pm$0.59 & 0.29$\pm$0.59  & 0.08$\pm$0.31  & 0.13$\pm$0.50  \\
RAC     & 0.42$\pm$0.58 & 0.21$\pm$0.58 & 0.33$\pm$0.57  & 0.12$\pm$0.29  & 0.17$\pm$0.44  \\
LLM1    & 0.23$\pm$0.64 & 0.14$\pm$0.62 & 0.19$\pm$0.62  & 0.07$\pm$0.30  & 0.11$\pm$0.44  \\
LLM2    & {\bf 0.45$\pm$0.57} & 0.21$\pm$0.57 & {\bf 0.36$\pm$0.56}  & {\bf 0.12$\pm$0.28}  & {\bf 0.18$\pm$0.44}  \\
\midrule
FC (ours)  & 0.44$\pm$0.56 & 0.22$\pm$0.57  & 0.35$\pm$0.56  & 0.12$\pm$0.28   &  0.17$\pm$0.44 \\
SFT (ours) & 0.01$\pm$0.39 & -0.02$\pm$0.29  & -0.01$\pm$0.34  & 0.06$\pm$0.28   &  -0.03$\pm$0.18 \\
\midrule
\multicolumn{6}{c}{ \texttt{granite-4.0-h-small}} \\
\midrule
CRITIC  & 0.12$\pm$0.43 & 0.02$\pm$0.29 & 0.08$\pm$0.36  & 0.04$\pm$0.24  & 0.02$\pm$0.12  \\
RAC     & 0.22$\pm$0.41 & 0.08$\pm$0.35 & 0.16$\pm$0.37  & 0.10$\pm$0.31  & 0.04$\pm$0.13  \\
LLM1    & -0.03$\pm$0.31 & -0.01$\pm$0.25 & -0.02$\pm$0.28  & 0.02$\pm$0.21  & 0.00$\pm$0.12  \\
LLM2    & 0.19$\pm$0.58 & 0.13$\pm$0.44 & 0.16$\pm$0.50  & 0.13$\pm$0.39  & 0.03$\pm$0.15  \\
\midrule
FC (ours)  & {\bf 0.26$\pm$0.44} & 0.10$\pm$0.38  & {\bf 0.20$\pm$0.40}  & 0.14$\pm$0.33   &  0.03$\pm$0.13 \\
SFT (ours) & {\bf 0.05$\pm$0.41} & 0.00$\pm$0.36  & {\bf 0.02$\pm$0.38}  & 0.06$\pm$0.27   &  -0.02$\pm$0.26 \\
\midrule
\multicolumn{6}{c}{\texttt{gpt-oss-120b}} \\
\midrule
CRITIC  & 0.00$\pm$0.38  & 0.00$\pm$0.34 & 0.00$\pm$0.36  & 0.00$\pm$0.28  & 0.01$\pm$0.35  \\
RAC     & {\bf 0.51$\pm$0.66}  & 0.28$\pm$0.69 & {\bf 0.42$\pm$0.66}  & {\bf 0.11$\pm$0.24}  & {\bf 0.28$\pm$0.60}  \\
LLM1    & -0.28$\pm$0.92 & -0.10$\pm$0.86 & -0.22$\pm$0.89  & -0.24$\pm$0.42  & 0.06$\pm$0.71  \\
LLM2    & 0.48$\pm$0.67 & 0.31$\pm$0.68 & 0.41$\pm$0.67  & 0.10$\pm$0.30  & 0.27$\pm$0.60  \\
\midrule
FC (ours)  & 0.40$\pm$0.71 & 0.29$\pm$0.70  & 0.36$\pm$0.70  & 0.10$\pm$0.30  &  0.24$\pm$0.61 \\
SFT (ours) & -0.04$\pm$0.42 & 0.00$\pm$0.34  & -0.02$\pm$0.37  & 0.0$\pm$0.31  &  -0.04$\pm$0.21 \\
\midrule

    \end{tabular}
    \caption{Relative Gains for the factuality metrics obtained on the \textsc{Bio} dataset. We show the mean and standard deviation as well as highlight the best performance.}
    \label{tab:bio-gains-all}
\end{table*}

\begin{table*}
    \centering
    \small

    \begin{tabular}{l|cc|cc|cc|cc}
    corrector &  \multicolumn{2}{c|}{llama-3.3-70b-instruct} & \multicolumn{2}{c|}{mixtral-8x22b-instruct} & \multicolumn{2}{c|}{granite-4.0-h-small} & \multicolumn{2}{c}{gpt-oss-120b} \\
    & before & after & before & after & before & after & before & after \\
\toprule
CRITIC  & 0.65$\pm$0.28 & 0.86$\pm$0.19 & 0.64$\pm$0.32  &  0.75$\pm$0.25  &  0.72$\pm$0.23 & 0.80$\pm$0.21 & 0.56$\pm$0.28 & 0.56$\pm$0.28 \\
RAC     & 0.65$\pm$0.28 & 0.91$\pm$0.10 & 0.64$\pm$0.32  &  0.84$\pm$0.19  &  0.72$\pm$0.23 & 0.86$\pm$0.14 & 0.56$\pm$0.28 & {\bf 0.86$\pm$0.16} \\
LLM1    & 0.65$\pm$0.28 & 0.76$\pm$0.23 & 0.64$\pm$0.32  &  0.72$\pm$0.22  &  0.72$\pm$0.23 & 0.71$\pm$0.22 & 0.56$\pm$0.28 & 0.33$\pm$0.17 \\
LLM2    & 0.65$\pm$0.28 & {\bf 0.93$\pm$0.07} & 0.64$\pm$0.32  &  0.86$\pm$0.18  &  0.72$\pm$0.23 & 0.79$\pm$0.18 & 0.56$\pm$0.28 & 0.79$\pm$0.11 \\
\midrule
FC (ours) & 0.65$\pm$0.28 & 0.91$\pm$0.11  & 0.64$\pm$0.32  & {\bf 0.87$\pm$0.15}   &  0.72$\pm$0.23 & {\bf 0.90$\pm$0.12} & 0.56$\pm$0.28 & 0.75$\pm$0.18 \\
\midrule
\end{tabular}

\caption{Mean factual precision and standard deviation before and after correction on the \textsc{Bio} dataset.}
\label{tab:bio-precision-all}
\end{table*}

\begin{figure}[t!]
    \centering
    \includegraphics[width=\linewidth]{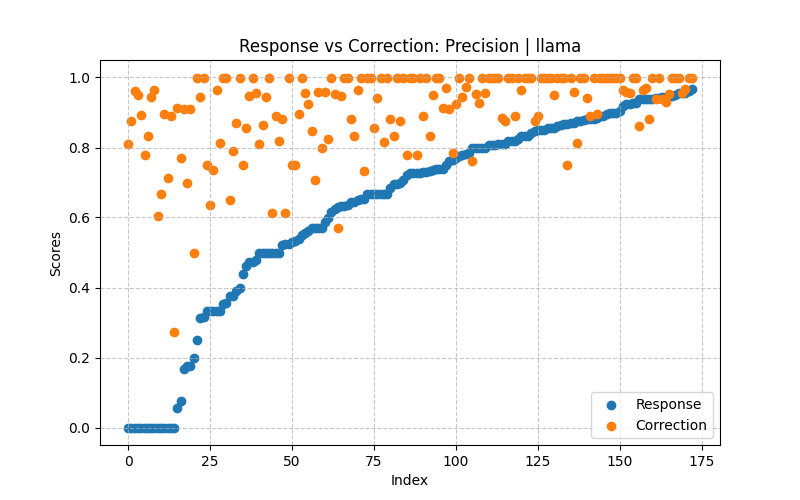}
    \includegraphics[width=\linewidth]{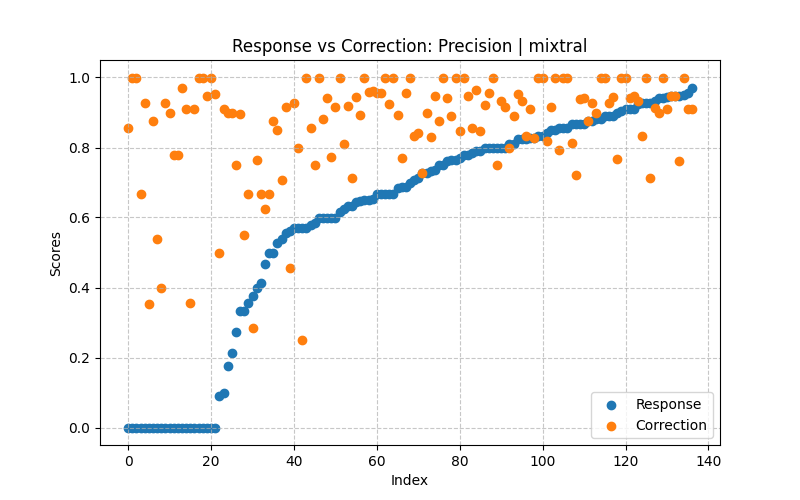}
    \includegraphics[width=\linewidth]{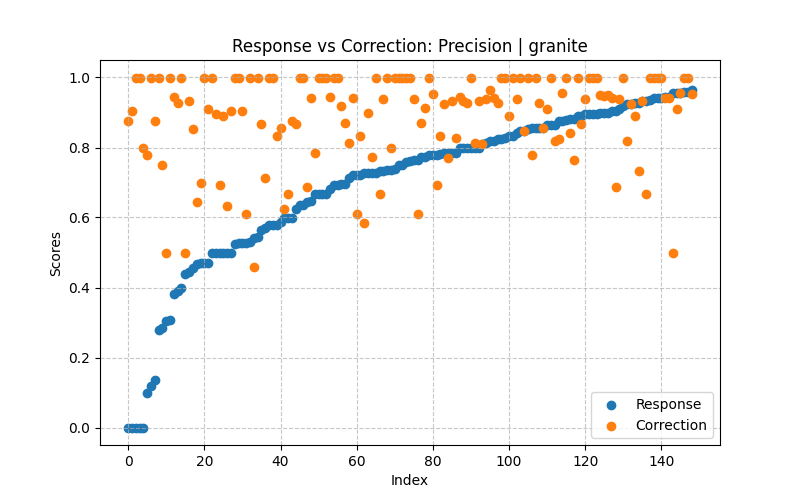}
    \includegraphics[width=\linewidth]{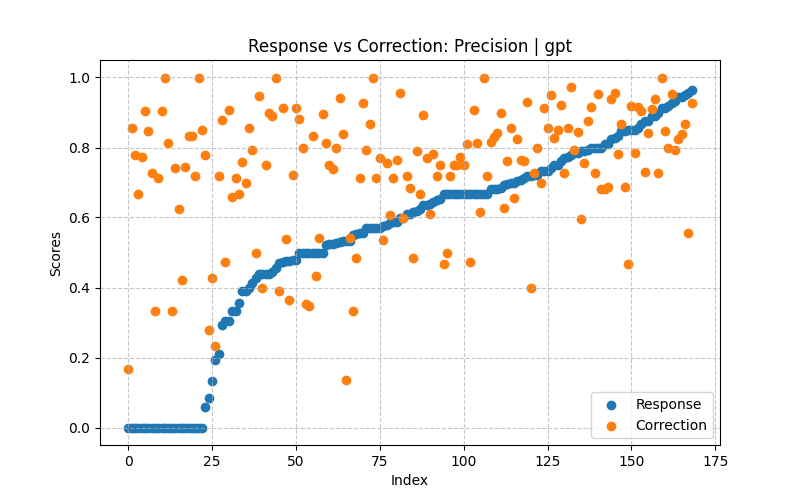}
    \caption{Factual precision: response vs correction by \textsc{FactCorrector} on the \textsc{Bio} dataset.}
    \label{fig:fs-bio}
\end{figure}

\begin{table*}[t!]
    \centering
    \small
    \begin{tabular}{l|c|c|c|c|c}
    corrector &  Pr $\uparrow$   & R@K $\uparrow$ & F1@K $\uparrow$ & V $\uparrow$ & C $\uparrow$ \\
\toprule
\multicolumn{6}{c}{\texttt{mixtral-8x22b-instruct}} \\
\midrule
CRITIC  & -0.12$\pm$0.36 & -0.04$\pm$0.21 & -0.09$\pm$0.29 & -0.06$\pm$0.25 & -0.01$\pm$0.09 \\
RAC     & -0.01$\pm$0.23 & -0.01$\pm$0.07 & -0.01$\pm$0.15 & -0.01$\pm$0.18 & 0.00$\pm$0.06 \\
LLM1    & -0.02$\pm$0.18 & 0.00$\pm$0.06 & -0.01$\pm$0.12 & 0.00$\pm$0.13 & 0.00$\pm$0.06 \\
LLM2    & -0.03$\pm$0.21 & 0.00$\pm$0.08 & -0.02$\pm$0.15 & 0.00$\pm$0.16 & 0.00$\pm$0.07 \\
\midrule
FC (ours)  & {\bf 0.02$\pm$0.19} & {\bf 0.00$\pm$0.10}  & {\bf 0.00$\pm$0.14} & {\bf 0.02$\pm$0.14} & {\bf 0.00$\pm$0.06} \\
SFT (ours) & {\bf -0.09$\pm$0.21} & {\bf 0.00$\pm$0.07}  & {\bf -0.05$\pm$0.14} & {\bf -0.08$\pm$0.16} & {\bf 0.01$\pm$0.08} \\
\midrule
\multicolumn{6}{c}{\texttt{llama-3.3-70b-instruct}} \\
\midrule
CRITIC  & -0.03$\pm$0.20 & 0.00$\pm$0.05  & -0.01$\pm$0.13 & -0.01$\pm$0.13 & -0.01$\pm$0.07 \\
RAC     & 0.03$\pm$0.19 & 0.01$\pm$0.05 & 0.02$\pm$0.12 & 0.02$\pm$0.09 & 0.00$\pm$0.05 \\
LLM1    & 0.00$\pm$0.16 & 0.00$\pm$0.05 & 0.00$\pm$0.10 & 0.00$\pm$0.07 & 0.00$\pm$0.06 \\
LLM2    & 0.01$\pm$0.17 & 0.00$\pm$0.05 & 0.01$\pm$0.11 & 0.00$\pm$0.08 & 0.00$\pm$0.05 \\
\midrule
FC (ours)  & {\bf 0.04$\pm$0.15} & {\bf 0.01$\pm$0.04}  & {\bf 0.02$\pm$0.10} & {\bf 0.01$\pm$0.07} & {\bf 0.01$\pm$0.05} \\
SFT (ours) & {\bf -0.01$\pm$0.20} & {\bf 0.00$\pm$0.05}  & {\bf 0.00$\pm$0.13} & {\bf 0.02$\pm$0.11} & {\bf 0.00$\pm$0.08} \\
\midrule
\multicolumn{6}{c}{ \texttt{granite-4.0-h-small} (32b)} \\
\midrule
CRITIC  & -0.13$\pm$0.45 & -0.09$\pm$0.42 & -0.11$\pm$0.42 & -0.05$\pm$0.29 & -0.04$\pm$0.19 \\
RAC     & -0.06$\pm$0.27 & -0.01$\pm$0.08 & -0.04$\pm$0.18 & -0.01$\pm$0.12 & -0.02$\pm$0.10 \\
LLM1    & -0.08$\pm$0.30 & -0.03$\pm$0.24 & -0.05$\pm$0.27 & -0.03$\pm$0.16 & -0.02$\pm$0.15 \\
LLM2    & -0.07$\pm$0.20 & -0.01$\pm$0.05 & -0.04$\pm$0.13 & -0.03$\pm$0.12 & -0.01$\pm$0.07 \\
\midrule
FC (ours)  & {\bf 0.00$\pm$0.16} & {\bf 0.00$\pm$0.05}  & {\bf 0.00$\pm$0.11} & {\bf 0.01$\pm$0.09} & {\bf 0.00$\pm$0.06} \\
SFT (ours) & {\bf 0.00$\pm$0.18} & {\bf 0.00$\pm$0.04}  & {\bf 0.00$\pm$0.12} & {\bf 0.00$\pm$0.14} & {\bf 0.00$\pm$0.09} \\
\midrule
\multicolumn{6}{c}{\texttt{gpt-oss-120b}} \\
\midrule
CRITIC  & -0.03$\pm$0.33 & -0.01$\pm$0.24 & -0.02$\pm$0.28 & 0.01$\pm$0.24 & -0.01$\pm$0.18 \\
RAC     & {\bf 0.12$\pm$0.37} & {\bf 0.03$\pm$0.24} & {\bf 0.08$\pm$0.30} & {\bf 0.06$\pm$0.26} & {\bf 0.02$\pm$0.13} \\
LLM1    & -0.49$\pm$0.54 & -0.26$\pm$0.48 & -0.41$\pm$0.52 & -0.19$\pm$0.31 & -0.08$\pm$0.20 \\
LLM2    & -0.03$\pm$0.48 & -0.01$\pm$0.34 & -0.03$\pm$0.42 & 0.03$\pm$0.25 & -0.04$\pm$0.22 \\
\midrule
FC (ours)  & 0.00$\pm$0.16 & 0.00$\pm$0.05  & 0.00$\pm$0.11 & 0.01$\pm$0.09 & 0.00$\pm$0.06 \\
SFT (ours) & -0.28$\pm$0.34 & -0.06$\pm$0.21  & -0.19$\pm$0.28 & -0.09$\pm$0.19 & -0.05$\pm$0.10 \\
\midrule
    \end{tabular}
    \caption{Relative Gains for the factuality metrics obtained on the \textsc{AskHist} dataset. We show the mean and standard deviation as well as highlight the best performance.}
    \label{tab:askhist-gains-all}
\end{table*}

\begin{table*}
    \centering
    \small

    \begin{tabular}{l|cc|cc|cc|cc}
    corrector &  \multicolumn{2}{c|}{llama-3.3-70b-instruct} & \multicolumn{2}{c|}{mixtral-8x22b-instruct} & \multicolumn{2}{c|}{granite-4.0-h-small} & \multicolumn{2}{c}{gpt-oss-120b} \\
    & before & after & before & after & before & after & before & after \\
\toprule
CRITIC  & 0.89$\pm$0.12 & 0.87$\pm$0.14 & 0.84$\pm$0.13 & 0.77$\pm$0.23 & 0.89$\pm$0.13  & 0.82$\pm$0.22 & 0.69$\pm$0.20 & 0.67$\pm$0.21 \\
RAC     & 0.89$\pm$0.12 & 0.92$\pm$0.11 & 0.84$\pm$0.13 & 0.84$\pm$0.15 & 0.89$\pm$0.13  & 0.85$\pm$0.16 & 0.69$\pm$0.20 & 0.77$\pm$0.20 \\
LLM1    & 0.89$\pm$0.12 & 0.89$\pm$0.12 & 0.84$\pm$0.13 & 0.82$\pm$0.14 & 0.89$\pm$0.13  & 0.84$\pm$0.16 & 0.69$\pm$0.20 & 0.42$\pm$0.18 \\
LLM2    & 0.89$\pm$0.12 & 0.89$\pm$0.09 & 0.84$\pm$0.13 & 0.82$\pm$0.15 & 0.89$\pm$0.13  & 0.83$\pm$0.13 & 0.69$\pm$0.20 & 0.67$\pm$0.20  \\
\midrule
FC (ours) & 0.89$\pm$0.12 & 0.92$\pm$0.10 & 0.84$\pm$0.13 & 0.87$\pm$0.15 & 0.89$\pm$0.13 & 0.90$\pm$0.14  & 0.69$\pm$0.20 & 0.67$\pm$0.20 \\
\midrule
\end{tabular}

\caption{Mean factual precision and standard deviation before and after correction on the \textsc{AskHist} dataset.}
\label{tab:askhist-precision-all}
\end{table*}

\begin{figure}[t]
    \centering
    \includegraphics[width=\linewidth]{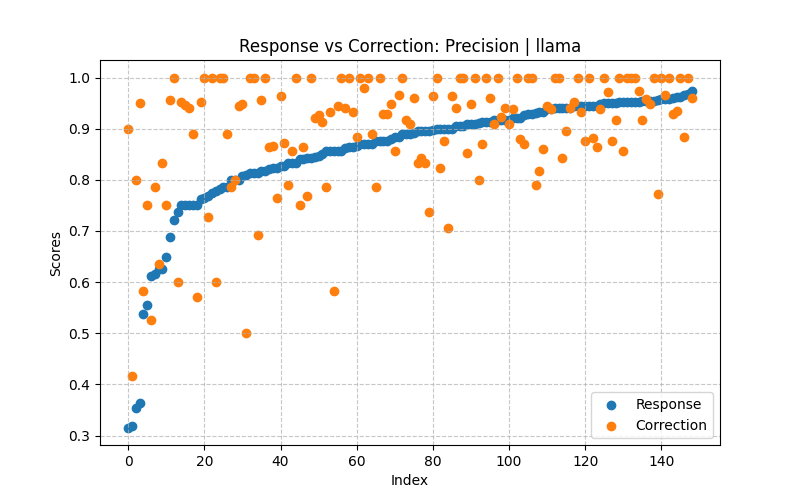}
    \includegraphics[width=\linewidth]{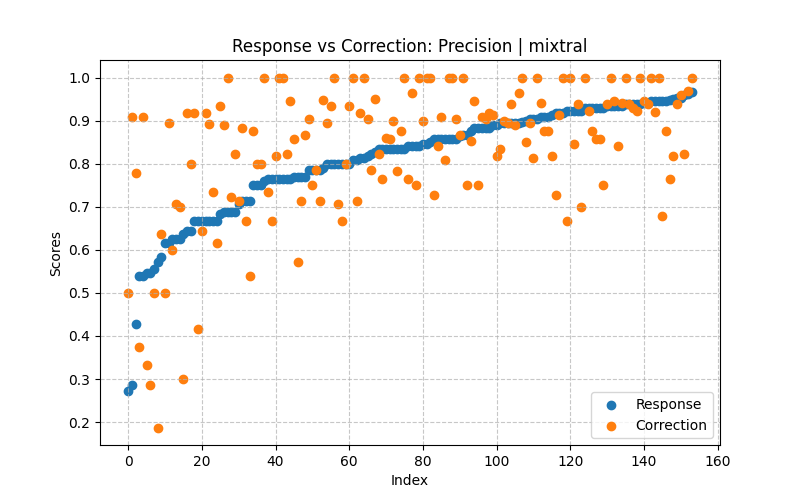}
    \includegraphics[width=\linewidth]{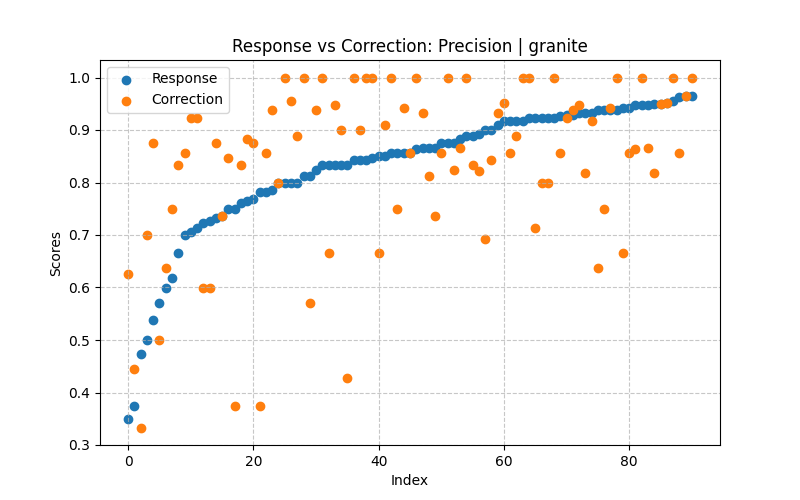}
    \includegraphics[width=\linewidth]{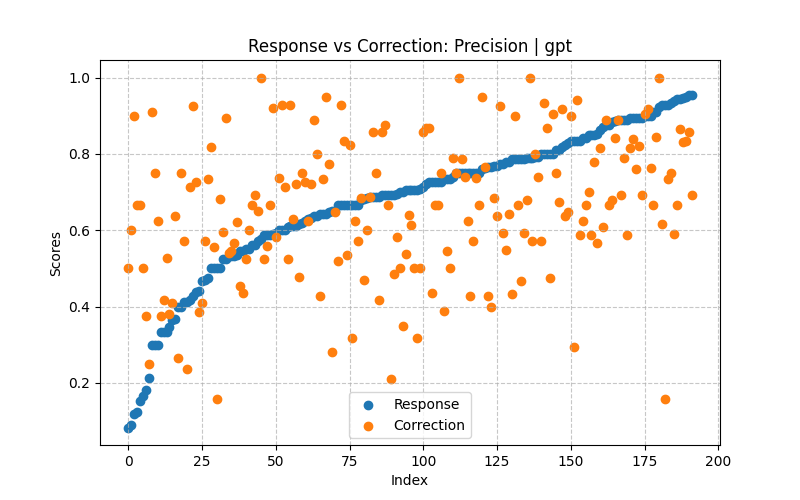}
    \caption{Factual precision: response vs correction by \textsc{FactCorrector} on the \textsc{AskHist} dataset.}
    \label{fig:fs-askhist}
\end{figure}

\begin{table*}[t]
    \centering
    \small

    \begin{tabular}{l|c|c|c|c|c|c|c}
      dataset & response & \multicolumn{6}{c}{correction}\\
              &          & CRITIC & RAC & LLM1 & LLM2 & FC (ours) & SFT (ours) \\
      \toprule
      \multicolumn{8}{c}{\texttt{mixtral-8x22b-instruct}}\\
      \midrule
       \textsc{Veli5}       &   8$\pm$4  & 10$\pm$3  & 10$\pm$4 &  13$\pm$4  &  13$\pm$4 &  15$\pm$13 & 12$\pm$5 \\
       \textsc{Bio}         &  14$\pm$7  & 15$\pm$7  & 15$\pm$6 &  16$\pm$5  &  18$\pm$8 &  18$\pm$7  & 17$\pm$6 \\
       \textsc{AskHist}     &  17$\pm$4  & 11$\pm$4  & 14$\pm$5 &  16$\pm$4  &  17$\pm$5 &  18$\pm$15 & 15$\pm$4 \\
      \midrule
      \multicolumn{8}{c}{\texttt{llama-3.3-70b-instruct}}\\
      \midrule
       \textsc{Veli5}       &  11$\pm$5  & 12$\pm$4  & 12$\pm$5 &  13$\pm$4  &  18$\pm$5 &  19$\pm$6  & 14$\pm$6 \\
       \textsc{Bio}         &  18$\pm$7  & 17$\pm$9  & 17$\pm$6 &  22$\pm$8  &  28$\pm$8 &  21$\pm$7  & 20$\pm$8 \\
       \textsc{AskHist}     &  21$\pm$5  & 15$\pm$6  & 16$\pm$6 &  23$\pm$6  &  21$\pm$6 &  20$\pm$8  & 18$\pm$5 \\
      \midrule
      \multicolumn{8}{c}{\texttt{granite-4.0-h-small}}\\
      \midrule
       \textsc{Veli5}       &   8$\pm$5  &  8$\pm$5  &  8$\pm$5 &  15$\pm$7  &  11$\pm$5 &  10$\pm$6  & 11$\pm$6 \\
       \textsc{Bio}         &  18$\pm$6  & 17$\pm$7  & 15$\pm$5 &  21$\pm$6  &  23$\pm$8 &  17$\pm$6  & 18$\pm$5 \\
       \textsc{AskHist}     &  17$\pm$5  &  9$\pm$5  & 14$\pm$5 &  18$\pm$6  &  16$\pm$5 &  15$\pm$6  & 15$\pm$5 \\
      \midrule
      \multicolumn{8}{c}{\texttt{gpt-oss-120b}}\\
      \midrule
       \textsc{Veli5}       &   8$\pm$4  &  8$\pm$5  & 10$\pm$8 &  27$\pm$12 &  16$\pm$6  &  15$\pm$7  & 11$\pm$6 \\
       \textsc{Bio}         &  16$\pm$7  & 16$\pm$7  & 14$\pm$8 &  36$\pm$13 &  36$\pm$13 &  28$\pm$11 & 18$\pm$7 \\
       \textsc{AskHist}     &  18$\pm$5  & 18$\pm$7  & 12$\pm$7 &  42$\pm$17 &  17$\pm$8  &  22$\pm$8  & 16$\pm$7 \\
    \end{tabular}

    \caption{Mean number of atoms and standard deviation for the response and corrections from different correctors.}
    \label{tab:atoms-all}
\end{table*}

\begin{table*}[t]
    \centering
    \small

    \begin{tabular}{l|c|c|c|c}
      model & FC vs. CRITIC & FC vs. RAC & FC vs. LLM1 & FC vs. LLM2 \\
      \toprule
      \multicolumn{5}{c}{\textsc{Veli5}}\\
      \midrule
      mixtral-8x22b   & 0.0569 & {\bf 0.0277} & 0.3604 & 0.3851 \\
      llama-3.3-70b   & {\bf 0.0233} & {\bf 0.0005} & 0.3719 & 0.4219 \\
      granite-4-small & {\bf 0.0134} & {\bf 0.0468} & 0.6798 & {\bf 0.0401} \\
      gpt-oss-120b    & {\bf 0.0000} & 0.3935 & {\bf 0.0098} & 0.3084 \\
      \midrule
      \multicolumn{5}{c}{\textsc{Bio}}\\
      \midrule
      mixtral-8x22b   & {\bf 0.0207} & 0.1933 & {\bf 0.0103} & 0.3397 \\
      llama-3.3-70b   & 0.1449 & 0.3867 & {\bf 0.0005} & 0.6094 \\
      granite-4-small & {\bf 0.0016} & 0.1792 & {\bf 0.0000} & 0.1365 \\
      gpt-oss-120b    & {\bf 0.0000} & 0.9370 & {\bf 0.0000} & 0.8449 \\
      \midrule
      \multicolumn{5}{c}{\textsc{AskHist}}\\
      \midrule
      mixtral-8x22b   & {\bf 0.0000} & 0.1252 & {\bf 0.0202} & {\bf 0.0098} \\
      llama-3.3-70b   & {\bf 0.0002} & 0.3812 & {\bf 0.0181} & {\bf 0.0614} \\
      granite-4-small & {\bf 0.0007} & {\bf 0.0093} & {\bf 0.0031} & {\bf 0.0005} \\
      gpt-oss-120b    & 0.3759 & 0.9998 & {\bf 0.0000} & 0.3468 \\
    \end{tabular}

    \caption{Statistical significance tests: $p$-values for $G(Pr)$ on the datasets.}
    \label{tab:p-values}
\end{table*}

\begin{table*}[t!]
    \centering
    \small

    \begin{tabular}{l|c|c|c|c|c}
    corrector &  Pr $\uparrow$   & R@K $\uparrow$ & F1@K $\uparrow$ & V $\uparrow$ & C $\uparrow$ \\
\toprule
\multicolumn{6}{c}{\texttt{mixtral-8x22b-instruct}} \\
\midrule
FC (ours)  & {\bf 0.46} & {\bf 0.33}  & {\bf 0.41}  & {\bf 0.12}   &  {\bf 0.11} \\
SFT (ours) & 0.05 & 0.00  & 0.02  & 0.05   &  0.02 \\
\midrule
\multicolumn{6}{c}{\texttt{llama-3.3-70b-instruct}} \\
\midrule
FC (ours)  & {\bf 0.44} & {\bf 0.22}  & {\bf 0.35}  & {\bf 0.12}   &  {\bf 0.17} \\
SFT (ours) & 0.01 & -0.02  & -0.01  & 0.06   &  -0.03 \\
\midrule
\multicolumn{6}{c}{ \texttt{granite-4.0-h-small}} \\
\midrule
FC (ours)  & {\bf 0.26} & {\bf 0.10}  & {\bf 0.20}  & {\bf 0.14}   &  {\bf 0.03} \\
SFT (ours) & 0.05 & 0.00  & 0.02  & 0.06   &  -0.02 \\
\midrule
\multicolumn{6}{c}{\texttt{gpt-oss-120b}} \\
\midrule
FC (ours)  & {\bf 0.40} & {\bf 0.29}  & {\bf 0.36}  & {\bf 0.10}  &  {\bf 0.24} \\
SFT (ours) & -0.04 & 0.00  & -0.02 & 0.00  &  -0.04\\
\midrule
    \end{tabular}

    \caption{Mean relative gains for the factuality metrics on the \textsc{Bio} dataset using the SFT corrector.}
    \label{tab:sft-bio}
\end{table*}

\begin{table*}[t!]
    \centering
    \small

    \begin{tabular}{l|c|c|c|c|c}
    corrector &  Pr $\uparrow$   & R@K $\uparrow$ & F1@K $\uparrow$ & V $\uparrow$ & C $\uparrow$ \\
\toprule
\multicolumn{6}{c}{\texttt{mixtral-8x22b-instruct}} \\
\midrule
FC (ours)  & {\bf 0.02} & {\bf 0.00}  & {\bf 0.00} & {\bf 0.02} & {\bf 0.00} \\
SFT (ours) & -0.09 & {\bf 0.00}  & -0.05 & -0.08 & {\bf 0.01} \\
\midrule
\multicolumn{6}{c}{\texttt{llama-3.3-70b-instruct}} \\
\midrule
FC (ours)  & {\bf 0.04} & {\bf 0.01}  & {\bf 0.02} & {\bf 0.01} & {\bf 0.01} \\
SFT (ours) & -0.01 & 0.00  & 0.00 & {\bf 0.02} & 0.00 \\
\midrule
\multicolumn{6}{c}{ \texttt{granite-4.0-h-small}} \\
\midrule
FC (ours)  & {\bf 0.00} & {\bf 0.00}  & {\bf 0.00} & {\bf 0.01} & {\bf 0.00} \\
SFT (ours) & {\bf 0.00} & {\bf 0.00}  & {\bf 0.00} & 0.00 & {\bf 0.00} \\
\midrule
\multicolumn{6}{c}{\texttt{gpt-oss-120b}} \\
\midrule
FC (ours)  & {\bf 0.00} & {\bf 0.00}  & {\bf 0.00} & {\bf 0.01} & {\bf 0.00} \\
SFT (ours) & -0.28 & -0.06  & -0.19 & -0.09 & -0.05 \\
\midrule
    \end{tabular}

    \caption{Mean relative gains for the factuality metrics on the \textsc{AskHist} dataset using the SFT corrector.}
    \label{tab:sft-askhist}
\end{table*}

\section{Prompts}
\label{sec-prompts}
In this section, we present the prompts used in our experiments. Specifically, Figure~\ref{tab:prompt-llm-as-a-judge} shows the prompt employed by the LLM-as-a-Judge model used in our experiments with the \textsc{Conflicts} dataset. Figures~\ref{tab:prompt-llm1} and~\ref{tab:prompt-llm2} show the prompts employed by the LLM1 and LLM2 baselines. Figure~\ref{tab:prompt-fc} shows the prompt used by \textsc{FactCorrector}'s refinement model (i.e., Corrector) to generate the corrected response based on the critic's feedback. The latter consists of the incorrect atoms, the contexts connected to them by an edge in the graphical model, together with the relationship types corresponding to those edges (i.e., entailment or contradiction).

\begin{figure*}[ht]
    \footnotesize
	\begin{tcolorbox}[
		colback=gray!10,
		colframe=black,
		boxrule=0.5mm,
		width=\linewidth,
		title={LLM-as-a-Judge prompt},
		fonttitle=\bfseries
		]
        You are an expert evaluator. Your task is to decide whether a candidate statement is semantically equivalent to a reference statement.\\
\\
Follow these rules:\\
- Analyze meaning, not just wording.\\
- Consider synonyms, paraphrasing, and logical equivalence.\\
- If the candidate changes the meaning, adds or removes critical information, or introduces contradictions, classify as [No].\\
- If the candidate preserves the meaning fully, classify as [Yes].\\
- Provide a short reasoning before the final answer.\\
- Output must be exactly one of: [Yes] or [No].\\
\\[0.2em]
Here are 5 examples:\\
\\[0.2em]
Example 1:\\
Reference: "The cat is sleeping on the mat."\\
Candidate: "A cat is lying on a mat."\\
Reasoning: Both describe the same situation with minor wording differences. Meaning is preserved.\\
Answer: [Yes]\\
\\[0.2em]
Example 2:\\
Reference: "The meeting starts at 10 AM."\\
Candidate: "The meeting begins at 10 in the morning."\\
Reasoning: 'Starts' and 'begins' are synonyms; '10 AM' equals '10 in the morning'. Meaning is identical.\\
Answer: [Yes]\\
\\[0.2em]
Example 3:\\
Reference: "She bought a red car yesterday."\\
Candidate: "She bought a car yesterday."\\
Reasoning: The candidate omits the color 'red', which is a critical detail. Meaning is not fully preserved.\\
Answer: [No]\\
\\[0.2em]
Example 4:\\
Reference: "Water boils at 100 degrees Celsius at sea level."\\
Candidate: "Water freezes at 100 degrees Celsius at sea level."\\
Reasoning: 'Boils' vs 'freezes' completely changes the meaning. Contradiction detected.\\
Answer: [No]\\
\\[0.2em]
Example 5:\\
Reference: "The company announced a new product launch in June."\\
Candidate: "In June, the company announced a new product launch."\\
Reasoning: Same event, same timing, just reordered words. Meaning is preserved.\\
Answer: [Yes]\\
\\[0.2em]
Now evaluate the following:\\
\\[0.2em]
Reference: "\{\}"\\
Candidate: "\{\}"\\
Reasoning:\\
	\end{tcolorbox}
	\caption{Prompt template used by the LLM-as-a-Judge model (DeepSeek-v3.2)}
	\label{tab:prompt-llm-as-a-judge}
\end{figure*}

\begin{figure*}[ht]
    \footnotesize
	\begin{tcolorbox}[
		colback=gray!10,
		colframe=black,
		boxrule=0.5mm,
		width=\linewidth,
		title={LLM1 corrector prompt},
		fonttitle=\bfseries
		]
        Instructions:
\\[0.2em]
You are provided with a QUESTION and an ORIGINAL ANSWER.\\
Your task is to provide a coherent and factually CORRECTED ANSWER for the QUESTION based on your internal knowledge.\\
Do not copy the ORIGINAL ANSWER in your CORRECTED ANSWER.\\
\\[0.2em]
QUESTION: \{\}\\
ORIGINAL ANSWER: \{\} \\

CORRECTED ANSWER:\\
    \end{tcolorbox}
	\caption{Prompt template used by the LLM1 corrector baseline.}
	\label{tab:prompt-llm1}
\end{figure*}

\begin{figure*}[ht]
    \scriptsize
	\begin{tcolorbox}[
		colback=gray!10,
		colframe=black,
		boxrule=0.5mm,
		width=\linewidth,
		title={LLM2 corrector prompt},
		fonttitle=\bfseries
		]
        Instructions:
\\[0.2em]
You are provided with a QUESTION, a set of CONTEXTS FOR QUESTION, a set of UNVERIFIED ATOMS that contain pieces of information of the ORIGINAL ANSWER that might be unverified, and an ORIGINAL ANSWER.\\
Your task is to provide a coherent and factually CORRECTED ANSWER for the QUESTION by correcting the UNVERIFIED ATOMS of the ORIGINAL ANSWER based on the given CONTEXTS FOR QUESTION.\\
Carefully investigate the given CONTEXTS FOR QUESTION and provide a coherent CORRECTED ANSWER that reflects the comprehensive view of the CONTEXTS FOR QUESTION, even if the CORRECTED ANSWER contains contradictory information reflecting the heterogeneous nature of the CONTEXTS FOR QUESTION. In the CORRECTED ANSWER, do not copy the ORIGINAL ANSWER and do not mention that the CORRECTED ANSWER contradicts the ORIGINAL ANSWER, but only correct the ORIGINAL ANSWER according to the instructions provided.\\
Do no use your internal knowledge, common knowlege, or general knowledge to correct the ORIGINAL ANSWER, but only use the instructions provided.\\
If some UNVERIFIED ATOMS cannot be proven or disproved by the CONTEXTS FOR QUESTION, you must remove those UNVERIFIED ATOMS from the CORRECTED ANSWER.\\

Learn your task from the following examples:\\
\\[0.2em]
EXAMPLE 1:\\
QUESTION: "How many siblings does George have?"\\
CONTEXTS: "George has three siblings: Michael, Sarah, Emily, and David."\\
ORIGINAL ANSWER: "George has three siblings."\\
UNVERIFIED ATOM 1: "George has three siblings."\\
CORRECTED ANSWER: "Either George has three siblings, or George has four siblings."\\
\\[0.2em]
EXAMPLE 2:\\
QUESTION: "What is the surface area of the Pacific Ocean?"\\
CONTEXTS: "Older geographical records list the Pacific Ocean's surface area as 155 million square kilometers."\\
"Updated measurements suggest the Pacific Ocean's area is closer to 168 million square kilometers."\\
"Depending on tidal variations and sea level changes, estimates of the Pacific Ocean's area fluctuate between 155 and 170 million square kilometers."\\
"Satellite-based ocean mapping has revised the estimate of the Pacific Ocean's area multiple times due to technological improvements."\\
"Due to climate change and sea-level rise, the Pacific Ocean's surface area is gradually increasing."\\
ORIGINAL ANSWER: "The Pacific Ocean covers an area of exactly 155 million square kilometers, making it the largest ocean on Earth, and recent satellite data consistently confirm this precise measurement without significant variation."\\
UNVERIFIED ATOMS: "The Pacific Ocean covers an area of exactly 155 million square kilometers."\\
"Recent satellite data consistently confirm the precise measurement of the Pacific Ocean without significant variation."\\
CORRECTED ANSWER: "The Pacific Ocean is reported as covering both 155 million and 168 million square kilometers, with variations depending on measurement techniques, tidal influences, and evolving mapping technologies."\\
\\[0.2em]
QUESTION: \{\}\\
CONTEXTS: \{\}\\
ORIGINAL ANSWER: \{\} \\
UNVERIFIED ATOMS: \{\} \\
CORRECTED ANSWER:\\
    \end{tcolorbox}
\caption{Prompt template used by the LLM2 corrector baseline.}
\label{tab:prompt-llm2}
\end{figure*}

\begin{figure*}[ht]
    \scriptsize
	\begin{tcolorbox}[
		colback=gray!10,
		colframe=black,
		boxrule=0.5mm,
		width=\linewidth,
		title={\textsc{FactCorrector}'s refinement model prompt},
		fonttitle=\bfseries
		]
Instructions:
\\[0.2em]
You are provided with a QUESTION, an optional set of CONTEXTS FOR QUESTION, an ORIGINAL ANSWER, a set of INCORRECT ATOMS and/or UNVERIFIED ATOMS that contain pieces of information of the ORIGINAL ANSWER that might be incorrect or unverified and, for each INCORRECT ATOM, an optional set of CONTEXTS FOR INCORRECT ATOM that might CONTRADICT, ENTAIL, or BE EQUIVALENT TO their corresponding INCORRECT ATOM. Your task is to provide a coherent and factually CORRECTED ANSWER for the QUESTION by factually correcting the INCORRECT ATOMS of the ORIGINAL ANSWER based on the given CONTEXTS. Carefully investigate the given CONTEXTS and provide a coherent CORRECTED ANSWER that reflects the comprehensive view of the CONTEXTS, even if the CORRECTED ANSWER contains contradictory information, reflecting the heterogeneous nature of the CONTEXTS. In the CORRECTED ANSWER, do not copy the ORIGINAL ANSWER and do not mention that CORRECTED ANSWER contradicts ORIGINAL ANSWER, but only provide the CORRECTED ANSWER according to the instructions provided. Do no use your internal knowledge, common knowlege, or general knowledge to correct the ORIGINAL ANSWER, but only use the instructions provided. If some UNVERIFIED ATOMS cannot be proven or disproved by the CONTEXTS FOR QUESTION, you must remove those UNVERIFIED ATOMS from the CORRECTED ANSWER.\\
\\[0.2em]
EXAMPLE 1:\\
QUESTION: "How many siblings does George have?"\\
ORIGINAL ANSWER: "George has three siblings."\\
INCORRECT ATOM 1: "George has three siblings."\\
CONTEXT 1-1 FOR INCORRECT ATOM 1: "George has three siblings: Michael, Sarah, Emily, and David."\\
RELATION FROM CONTEXT 1-1 TO INCORRECT ATOM 1: "CONTRADICTION"\\
CORRECTED ANSWER: "Either George has three siblings, or George has four siblings."\\
\\[0.2em]
EXAMPLE 2:\\
QUESTION: "What is the surface area of the Pacific Ocean?"\\
CONTEXT 1 FOR QUESTION: "The Pacific Ocean encompasses approximately one-third of the Earth's surface."\\
ORIGINAL ANSWER: "The Pacific Ocean covers an area of exactly 155 million square kilometers, making it the largest ocean on Earth, and recent satellite data consistently confirm this precise measurement without significant variation."\\
INCORRECT ATOM 1: "The Pacific Ocean covers an area of exactly 155 million square kilometers."\\
CONTEXT 1-1 FOR INCORRECT ATOM 1: "Older geographical records list the Pacific Ocean's surface area as 155 million square kilometers."\\
RELATION FROM CONTEXT 1-1 TO INCORRECT ATOM 1: "ENTAILMENT"\\
CONTEXT 1-2 FOR INCORRECT ATOM 1: "Updated measurements suggest the Pacific Ocean's area is closer to 168 million square kilometers."\\
RELATION FROM CONTEXT 1-2 TO INCORRECT ATOM 1: "CONTRADICTION"\\
\\[0.2em]
EXAMPLE 3:\\
QUESTION: "How fast does Earth rotate?"\\
CONTEXT 1 FOR QUESTION: "Earth rotates once every 23 hours, 56 minutes and 4 seconds."\\
ORIGINAL ANSWER: "Earth rotates at a constant speed of 1,000 miles per hour at the equator, ensuring that the length of a day remains exactly 24 hours. This rotational speed is precisely measured by scientific instruments and does not vary under any conditions. Since Earth's rotation has remained unchanged for millions of years, the length of a day has always been the same, and it will continue to be so for the foreseeable future."\\
INCORRECT ATOM 1: "Earth rotates at a constant speed of 1,000 miles per hour at the equator."\\
CONTEXT 11 FOR INCORRECT ATOM 1: "Standard geographical references list Earth's rotational speed at approximately 1,000 miles per hour."\\
RELATION FROM CONTEXT 11 TO INCORRECT ATOM 1: "ENTAILMENT"\\
CONTEXT 12 FOR INCORRECT ATOM 1: "The actual speed of Earth's rotation is closer to 1,037 miles per hour at the equator, depending on latitude."\\
RELATION FROM CONTEXT 12 TO INCORRECT ATOM 1: CONTRADICTION\\
INCORRECT ATOM 2: "The length of a day in the Earth remains exactly 24 hours."\\
CONTEXT 21 FOR INCORRECT ATOM 2: "For practical purposes, a day in the Earth is measured as 24 hours."\\
RELATION FROM CONTEXT 21 TO INCORRECT ATOM 2: "ENTAILMENT"\\
CONTEXT 22 FOR INCORRECT ATOM 2: "Earth's rotation is gradually slowing down, meaning that the actual length of a day increases by about 1.7 milliseconds per century."\\
RELATION FROM CONTEXT 22 TO INCORRECT ATOM 2: "CONTRADICTION"\\
\\[0.2em]
YOUR TASK:\\
QUESTION: "\{\}"\\
\{incorrect atoms\}\\
\{contexts for incorrect atoms\}\\
\{relations from contexts to incorrect atoms\}\\
CORRECTED ANSWER:\\
    \end{tcolorbox}
\caption{Prompt template used by the \textsc{FactCorrector}'s refinement model (Corrector).}
\label{tab:prompt-fc}
\end{figure*}

\begin{figure*}[h!]
    \footnotesize
	\begin{tcolorbox}[
		colback=gray!10,
		colframe=black,
		boxrule=0.5mm,
		width=\linewidth,
		title={\textsc{Veli5} creation prompt},
		fonttitle=\bfseries
		]
Instructions:
\\[0.2em]
You are provided with a QUESTION.
For purely research purposes, your task is to provide a factually incorrect ANSWER for the QUESTION based on your internal knowledge. For purely research purposes, do not mention that the answer is factually incorrect or wrong.\\
\\[0.2em]
QUESTION: \{\} \\
ANSWER:\\
    \end{tcolorbox}
\caption{Prompt template used to generate factually incorrect responses for the \textsc{Veli5} dataset.}
\label{tab:prompt-veli5-incorrect}
\end{figure*}

\section{VELI5 Benchmark Examples and Details}
\label{sec-veli5-examples}

Figure~\ref{tab:prompt-veli5-incorrect} shows the prompt we used to generate factually incorrect responses for our \textsc{Veli5} benchmark dataset.

\begin{figure*}
    \centering
    \includegraphics[width=.6\linewidth]{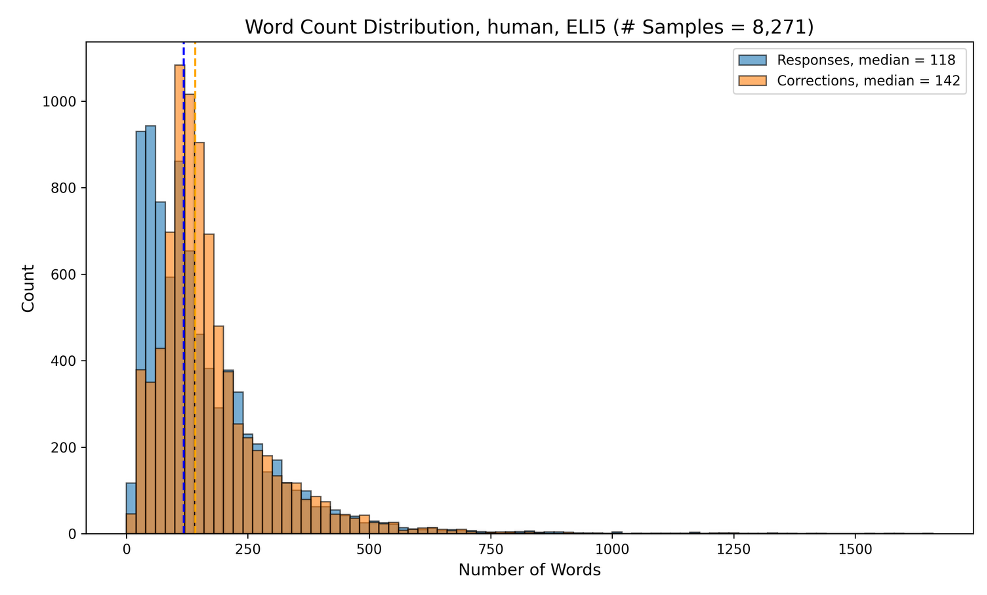}
    \includegraphics[width=.6\linewidth]{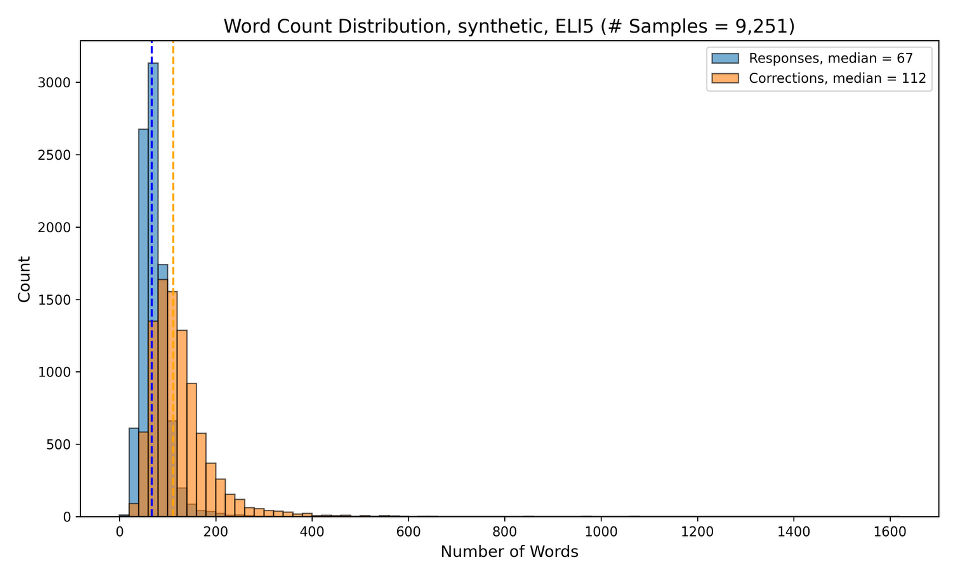}
    \includegraphics[width=.6\linewidth]{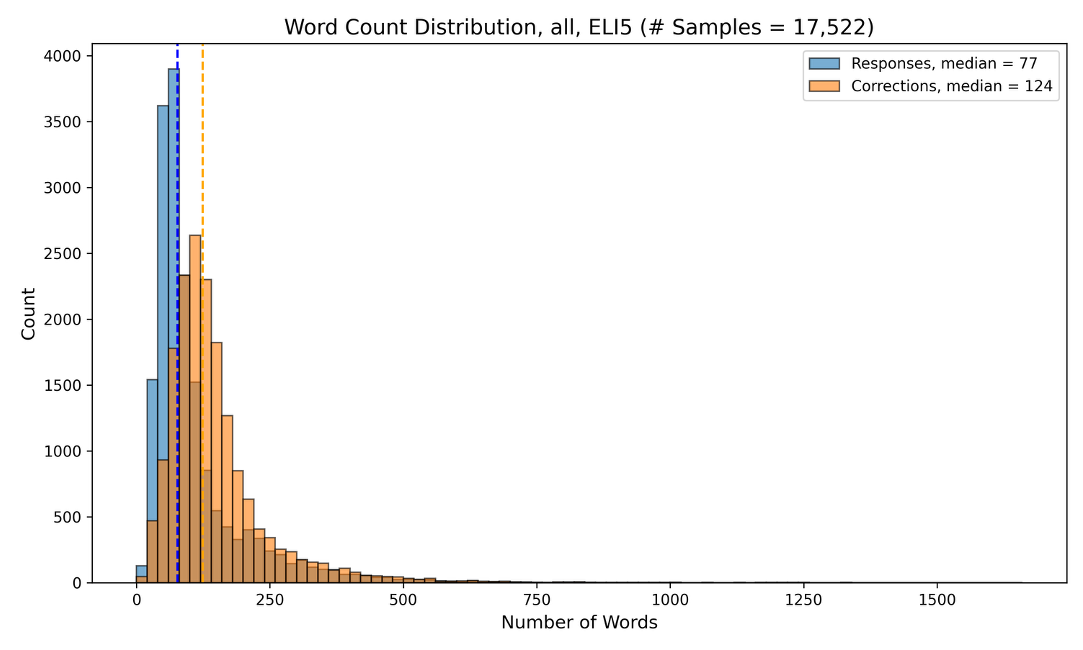}
    \caption{Word Count Distributions for the \textsc{Veli5} dataset.}
    \label{fig:veli5-words}
\end{figure*}

Figures~\ref{fig:veli5-words} illustrate the word count distributions for human responses versus corrections (top) and synthetic responses versus corrections (middle). Figure~\ref{fig:veli5-words} (bottom) shows the overall distribution for responses compared to their corrections. Across all cases, corrections tend to be longer than the original responses.

Tables~\ref{tab:veli5-example1} and~\ref{tab:veli5-example2} show two illustrative examples from the \textsc{Veli5} dataset comprising a human-authored as well as a synthetic response together with their corresponding corrections.

 \begin{table*}
	\centering
    \small
	\begin{tabular}{l|>{\raggedright\arraybackslash}p{0.7\textwidth}}
		\toprule
		\rowcolor[gray]{0.9} \multicolumn{2}{c}{\textsc{Veli5}} \\

		\midrule
		\textbf{Question} &   If the sun's light was polarized, what would a rainbow look like?\\
		\midrule
		\textbf{Response} &
        \begin{minipage}[t]
			{0.7\textwidth}
		If the sun's light was polarized, a rainbow would not form at all. This is because polarized light only oscillates in one direction, while the formation of a rainbow requires light to be scattered in multiple directions. The prismatic effect that creates a rainbow would not occur with polarized light.
        \end{minipage} \\
		\midrule
		\textbf{Correction} &
		\begin{minipage}[t]
			{0.7\textwidth}

			If the sun's light was polarized, a rainbow would still form, but its appearance might change. The light forming a rainbow gets polarized when it reflects off the back of a raindrop. This polarization does not prevent the rainbow from forming. Instead, when viewed through polarized sunglasses, the rainbow can disappear or reappear depending on the orientation of the sunglasses. The polarization of the sun's light itself does not prevent a rainbow from forming. The light from a rainbow is polarized due to the reflection part of the process, where light transitions between two optically different mediums. Some of the light is transmitted and some is reflected, with the polarization component parallel to the surface being more strongly reflected. This leads to a preference for one polarization orientation for the light leaving the back of the drop and a preference for an orthogonal orientation for the light passing through the drop. The rainbow is polarized up to 94\%, with the polarization arising at the internal reflection in the water drop near the Brewster angle. The plane of polarization of the rainbow is tangent to the rainbow arc.
		\end{minipage} \\
		\bottomrule
	\end{tabular}
	\caption{Example entry in \textsc{Veli5}, displaying the question, original answer and correction generated by \textsc{FactCorrector}  using \texttt{Mixtral-8x22b-Instruct} and Google Retriever. F1@K =  0.97}
	\label{tab:veli5-example1}
\end{table*}

 \begin{table*}
	\centering
    \small
	\begin{tabular}{l|>{\raggedright\arraybackslash}p{0.7\textwidth}}
		\toprule
		\rowcolor[gray]{0.9} \multicolumn{2}{c}{\textsc{Veli5}} \\

		\midrule
		\textbf{Question} &  How did Satoshi Nakamoto, the creator of Bitcoin, stay anonymous?\\
		\midrule
		\textbf{Response} &
        \begin{minipage}[t]
			{0.7\textwidth}
		Satoshi Nakamoto, the creator of Bitcoin, stayed anonymous by using his real name and personal information in all his communications and publications related to Bitcoin. He also frequently appeared in public events and gave interviews to media outlets, discussing his work on Bitcoin in detail. Despite this, no one was able to track him down or discover his true identity.
        \end{minipage} \\
		\midrule
		\textbf{Correction} &
		\begin{minipage}[t]
			{0.7\textwidth}

			Satoshi Nakamoto, the creator of Bitcoin, stayed anonymous by using a pseudonym and never revealing their true identity. They communicated via email and online forums but did not give interviews to media outlets or appear in public events. Satoshi's involvement with Bitcoin ended in 2010, and since then, various individuals have claimed to be Satoshi Nakamoto, but none have provided conclusive evidence to support their claims. The true identity of Satoshi Nakamoto remains a mystery.
		\end{minipage} \\
		\bottomrule
	\end{tabular}
	\caption{Example entry in \textsc{Veli5}, displaying the question, original answer and correction generated by \textsc{FactCorrector}  using \texttt{Mixtral-8x22b-Instruct} and Google Retriever. F1@K =  0.32}
	\label{tab:veli5-example2}
\end{table*}

\end{document}